\documentclass[journal,twoside]{IEEEtran}

\usepackage{amsmath,amssymb,amsfonts}
\usepackage{algorithmic}
\usepackage{algorithm}
\usepackage{graphicx}
\usepackage{textcomp}
\usepackage{xcolor}
\usepackage{booktabs}
\usepackage{cite}
\usepackage{bm}
\usepackage{url}
\usepackage{tikz}
\usetikzlibrary{arrows.meta, positioning, shapes.geometric, shapes.misc, shapes.symbols, fit, backgrounds}
\usepackage{tabularx}
\usepackage{multirow}

\begin{document}

\title{Energy-Efficient Hierarchical Federated Anomaly Detection for the Internet of Underwater Things via Selective Cooperative Aggregation}

\author{Kenechi Omeke,~\IEEEmembership{Member,~IEEE}, Michael Mollel,~\IEEEmembership{Member,~IEEE}, Lei Zhang,~\IEEEmembership{Senior Member,~IEEE}, Qammer H. Abbasi,~\IEEEmembership{Senior Member,~IEEE}, and Muhammad Ali Imran,~\IEEEmembership{Fellow,~IEEE}
\thanks{The authors are with the James Watt School of Engineering, University of Glasgow, U.K. (E-mail: \{Kenechi.Omeke, Michael.Mollel, Lei.Zhang, Qammer.Abbasi, Muhammad.Imran\}@glasgow.ac.uk). \\
Corresponding author: Kenechi Omeke (kenechiomeke@gmail.com).}}

\markboth{IEEE Internet of Things Journal,~Vol.~XX, No.~X, 2026}{Omeke \MakeLowercase{\textit{et al.}}: Selective Cooperative HFL for IoUT}

\maketitle

% --- Abstract ---
\begin{abstract}
Anomaly detection is a core service in the Internet of Underwater Things, yet training accurate distributed models underwater is difficult because acoustic links are low-bandwidth, energy-intensive, and often unable to support direct sensor-to-surface communication. Standard flat federated learning therefore faces two coupled limitations in underwater deployments: expensive long-range transmissions and reduced participation when only a subset of sensors can reach the gateway. This paper proposes an energy-efficient hierarchical federated learning framework for underwater anomaly detection based on three components: feasibility-aware sensor-to-fog association, compressed model-update transmission, and selective cooperative aggregation among fog nodes. The proposed three-tier architecture localises most communication within short-range clusters while activating fog-to-fog exchange only when smaller clusters can benefit from nearby larger neighbours. A physics-grounded underwater acoustic model is used to evaluate detection quality, communication energy, and network participation jointly. In large synthetic deployments, only about 48\% of sensors can directly reach the gateway in the 200-sensor case, whereas hierarchical learning preserves full participation through feasible fog paths. Selective cooperation matches the detection accuracy of always-on inter-fog exchange while reducing its energy by 31--33\%, and compressed uploads reduce total energy by 71--95\% in matched sensitivity tests. Experiments on three real benchmarks further show that low-overhead hierarchical methods remain competitive in detection quality, while flat federated learning defines the minimum-energy operating point. These results provide practical design guidance for underwater deployments operating under severe acoustic communication constraints.
\end{abstract}


\section{Introduction}
\label{sec:introduction}

\IEEEPARstart{T}{he} \emph{Internet of Underwater Things} (IoUT) extends the terrestrial Internet of Things (IoT) into subsea environments, enabling persistent ocean observation, offshore infrastructure monitoring, environmental protection, and autonomous underwater operations~\cite{akyildiz2005underwater,qiu2019underwater,jouhari2019underwater,adam2024iout_security}. As these deployments grow in scale and autonomy, they increasingly require in-network intelligence rather than simple data collection. Among the most important of these intelligent services is \emph{anomaly detection}, which enables early discovery of sensor faults, abnormal environmental events, and malicious behaviour from distributed multivariate time series.

Delivering such intelligence underwater is far more difficult than in terrestrial IoT. IoUT nodes communicate predominantly through acoustic links whose low bandwidth, long propagation delay, high bit-error rate, and severe attenuation make frequent long-range communication expensive and unreliable~\cite{stojanovic2009underwater,zeng2020survey_uwan}. Combined with limited onboard energy and the practical difficulty of battery replacement or maintenance, these channel characteristics make raw-data centralisation unattractive for learning tasks.

\emph{Federated learning}~(FL) is therefore a natural candidate for IoUT because it avoids transferring raw measurements and instead exchanges model updates~\cite{mcmahan2017communication,li2020fl_challenges_survey}. However, conventional star-topology FL still assumes that every sensor can repeatedly communicate with a central coordinator. Underwater, this assumption is problematic for two reasons. First, direct sensor-to-surface transmissions are often the most energy-expensive links in the network. Second, some sensors may not be able to reach the surface gateway at all under realistic acoustic constraints. As a result, a flat FL system can appear highly efficient while training on only the directly reachable subset of the deployment.

\emph{Hierarchical federated learning}~(HFL) offers a promising alternative by inserting fog or edge aggregators between the sensors and the global server~\cite{liu2020client,luo2020hfel_twc,wang2022feduc}. In IoUT, this hierarchy is physically meaningful: clusters of seafloor sensors can upload to nearby mid-water aggregators, such as autonomous underwater vehicles (AUVs) or anchored relays, which then communicate aggregated models to a surface gateway. This architecture reduces the burden of long-range transmissions, but it introduces a second design challenge. If fog-to-fog model exchange is enabled indiscriminately, the resulting inter-fog traffic may erase a large part of the expected energy gain.

The central problem addressed in this paper is therefore not simply whether HFL is better than flat FL, but \emph{how hierarchical cooperation should be designed for underwater learning under realistic acoustic constraints}. A useful IoUT learning framework must answer three practical questions at once: which sensors can actually participate, how much communication energy can be reduced through compressed uplinks, and when cooperative aggregation among fog nodes is beneficial enough to justify its additional cost. Existing IoUT studies on FL do not address these questions jointly and generally stop short of a physics-grounded evaluation of learning quality, communication energy, and effective network participation~\cite{pei2023fediout,popli2025fl_underwater_drones,shaheen2024fl_iout}.

% Ensure these are in your preamble:
% \usetikzlibrary{positioning, shapes.symbols, arrows.meta}

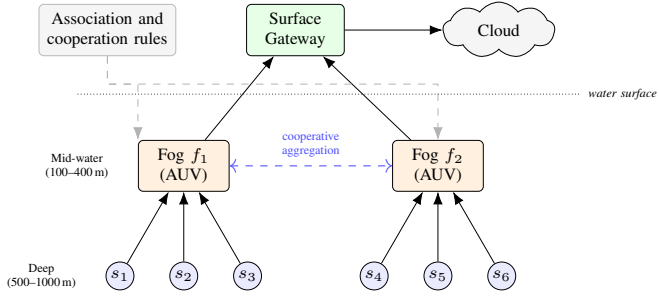
\begin{figure}[t]
\centering
\resizebox{\columnwidth}{!}{%
\begin{tikzpicture}[
  node distance=4mm and 7mm,
  every node/.style={font=\scriptsize, align=center},
  sens/.style={circle, draw, fill=blue!8, minimum size=4mm, inner sep=0pt},
  fog/.style={rectangle, draw, fill=orange!12, rounded corners=2pt, 
              minimum width=13mm, minimum height=6.5mm},
  surf/.style={rectangle, draw, fill=green!10, rounded corners=2pt, 
               minimum width=14mm, minimum height=7mm},
  % Renamed to 'cloudstyle' to avoid conflict with the 'cloud' shape name
  cloudstyle/.style={cloud, draw, fill=gray!8, cloud puffs=9, cloud puff arc=110, 
                aspect=2, minimum width=16mm, minimum height=8mm},
  arr/.style={-{Latex[length=1.8mm]}, line width=0.35pt},
  coop/.style={<->, dashed, line width=0.5pt, blue!60},
  rulebox/.style={rectangle, draw=gray!55, fill=gray!8, rounded corners=2pt, 
                minimum width=16mm, minimum height=7mm, font=\scriptsize}
]

% ----- Sensor layer -----
\node[sens] (s1) {$s_1$};
\node[sens, right=5mm of s1] (s2) {$s_2$};
\node[sens, right=5mm of s2] (s3) {$s_3$};
\node[sens, right=14mm of s3] (s4) {$s_4$};
\node[sens, right=5mm of s4] (s5) {$s_5$};
\node[sens, right=5mm of s5] (s6) {$s_6$};

% ----- Fog layer -----
\node[fog, above=10mm of s2] (f1) {Fog $f_1$\\(AUV)};
\node[fog, above=10mm of s5] (f2) {Fog $f_2$\\(AUV)};

% ----- Surface -----
\node[surf, above=12mm of f1, xshift=16mm] (gw) {Surface\\Gateway};
\node[cloudstyle, right=14mm of gw] (cld) {Cloud};

% ----- Association / cooperation rules -----
\node[rulebox, left=10mm of gw] (rules) {Association and\\cooperation rules};

% ----- Links: sensor -> fog -----
\draw[arr] (s1) -- (f1);
\draw[arr] (s2) -- (f1);
\draw[arr] (s3) -- (f1);
\draw[arr] (s4) -- (f2);
\draw[arr] (s5) -- (f2);
\draw[arr] (s6) -- (f2);

% ----- Fog cooperation -----
\draw[coop] (f1) -- node[above, font=\tiny, text=blue!70]{cooperative\\aggregation} (f2);

% ----- Fog -> surface -----
\draw[arr] (f1) -- (gw);
\draw[arr] (f2) -- (gw);

% ----- Surface -> cloud -----
\draw[arr] (gw) -- (cld);

% ----- Rule arrows (dashed) -----
\draw[arr, dashed, gray!60] (rules.south) -- ++(0,-4mm) -| (f1.north west);
\draw[arr, dashed, gray!60] (rules.south) -- ++(0,-4mm) -| (f2.north);

% ----- Layer labels -----
\draw[densely dotted, line width=0.3pt] ([xshift=-4mm]s1.west |- f1.south) ++(0,14mm) -- ++(72mm,0) 
  node[right, font=\tiny]{\textit{water surface}};
\node[font=\tiny, anchor=east] at ([xshift=-3mm]s1.west) {Deep\\(500--1000\,m)};
\node[font=\tiny, anchor=east] at ([xshift=-3mm]f1.west) {Mid-water\\(100--400\,m)};

\end{tikzpicture}%
}% end resizebox
\vspace{-1mm}
\caption{Proposed three-tier HFL architecture for IoUT with feasibility-aware association, compressed uplinks, and selective fog cooperation.}
\label{fig:hfl_concept}
\vspace{-3mm}
\end{figure}

In response, this paper proposes a three-tier \textbf{hierarchical federated anomaly-detection framework with selective cooperative aggregation} for IoUT, illustrated in Fig.~\ref{fig:hfl_concept}. The proposed framework combines feasibility-aware sensor-to-fog association, compressed model-update transmission, and selective fog-to-fog cooperation so that the communication architecture remains energy-efficient without sacrificing the ability to learn from the full underwater deployment. Equally importantly, the framework is constructed to evaluate the learning system in the way an IoUT designer actually needs to see it: not only in terms of detection accuracy and energy consumption, but also in terms of how much of the network can effectively participate in training. Its central contribution is to make these feasibility constraints explicit within a participation-aware evaluation framework, allowing the flat-versus-hierarchical design trade-off to be quantified under realistic IoUT conditions.

The main contributions of the paper are as follows:
\begin{enumerate}
  \item \textbf{Participation-aware hierarchical IoUT learning framework.}
    We formulate a three-tier underwater learning system (sensor $\rightarrow$ fog $\rightarrow$ surface) with a capped-source-level acoustic channel model, feasible communication graph, and explicit participation accounting, enabling realistic evaluation of who can train, at what communication cost, and through which underwater paths.
  \item \textbf{Selective cooperative aggregation with compressed uplinks.}
    We introduce a lightweight inter-fog cooperation strategy that activates only when smaller fog clusters can benefit from nearby larger neighbours, and we combine it with compressed sensor uploads to make hierarchical training practical over acoustic links.
  \item \textbf{Comprehensive evaluation and IoUT design guidance.}
    Through synthetic experiments and real-data evaluation on the Server Machine Dataset (SMD), Soil Moisture Active Passive (SMAP), and Mars Science Laboratory (MSL) benchmarks, we quantify the trade-offs among participation, detection accuracy, and energy consumption, and derive practical guidance for when flat FL is sufficient and when hierarchy becomes necessary to retain feasible participation.
\end{enumerate}

\smallskip
\noindent\textbf{Organisation.}
Section~\ref{sec:related_work} surveys related work.
Section~\ref{sec:system_model} presents the system and channel model.
Section~\ref{sec:problem_formulation} formalises the optimisation problem and decision space.
Section~\ref{sec:framework} details the proposed hierarchical FL framework.
Section~\ref{sec:evaluation} presents implementation details and performance results, and
Section~\ref{sec:conclusion} concludes the paper.

\section{Related Work}
\label{sec:related_work}

This section reviews three intersecting threads: IoUT architectures and anomaly detection; federated and hierarchical FL; and communication- and energy-efficient FL. Adaptive orchestration, including RL-based approaches, is noted only briefly as background because the main paper studies transparent deterministic association and cooperation rules rather than learned controllers.
It then synthesises the gaps that motivate the proposed hierarchical IoUT learning framework.

% ------------------------------------------------------------------
\subsection{IoUT Architectures, Anomaly Detection, and Edge Intelligence}
\label{subsec:rw_iout}

IoUT extends underwater sensor networks and AUV systems toward persistent, data-driven subsea cyber-physical systems~\cite{akyildiz2005underwater,qiu2019underwater,jouhari2019underwater}.
Acoustic channels impose severe bandwidth, delay, and energy constraints that complicate reliable multi-hop networking and make frequent control signalling costly~\cite{stojanovic2009underwater,zeng2020survey_uwan,chitre2008underwater}.
Edge and fog computing have been proposed to push intelligence closer to the data sources, reducing dependence on surface-level processing~\cite{xiao2020edge_iout,stewart2025iout_oil_gas}.

Anomaly detection in resource-constrained sensing systems has evolved from statistical and distance-based methods to deep representation learning, including recurrent sequence models~\cite{malhotra2016lstmencdec}, one-class objectives such as Deep~SVDD~\cite{ruff2018deepsvdd}, memory-augmented autoencoders~\cite{gong2019memae}, and more recent transformer-based approaches~\cite{pang2021deep_ad_survey,ruff2021unifying_ad}.
In IoUT specifically, anomaly detection must contend with non-stationarity (environmental drift), missing data from intermittent links, and significant device heterogeneity---motivating adaptive, distributed learning strategies rather than static thresholds or centralised models.

% ------------------------------------------------------------------
\subsection{Federated and Hierarchical Federated Learning}
\label{subsec:rw_fl_hfl}

FL trains models collaboratively without centralising raw data, typically via alternating local training and server-side aggregation~\cite{mcmahan2017communication}. Canonical algorithms include Federated Averaging (FedAvg)~\cite{mcmahan2017communication}, Federated Proximal (FedProx)~\cite{li2020fedprox}, and SCAFFOLD (Stochastic Controlled Averaging for Federated Learning)~\cite{karimireddy2020scaffold}.
Key challenges under heterogeneous and communication-constrained settings include client drift from non-independent and identically distributed (non-IID) data, straggler effects~\cite{wang2019adaptive}, and communication overhead.
FL surveys for IoT document these challenges extensively~\cite{nguyen2021fliotsurvey,li2020fl_challenges_survey,khan2021fl_iot_survey}.

HFL introduces one or more intermediate aggregation tiers (client~$\to$~edge/fog~$\to$~cloud) to reduce long-range update frequency~\cite{liu2020client,abad2019hflcellular,luo2020hfel_twc}. Recent advances include Hierarchical Federated Edge Learning (HFEL)~\cite{luo2020hfel_twc}, the Federated Unified Clustering approach (FedUC)~\cite{wang2022feduc}, adaptive asynchronous schemes~\cite{liu2023async_fl_mobile}, and resource-efficient edge assignment~\cite{mhaisen2022hfl_topology}. Clustered FL also partitions clients into groups that better match statistical heterogeneity~\cite{ghosh2020ifca}, while decentralised FL replaces the central server with peer-to-peer consensus~\cite{lian2017decentralized_sgd}.
Despite conceptual appeal for underwater multi-hop topologies, these frameworks universally assume terrestrial networking abstractions---high-rate links, sub-millisecond round-trip times, and reliable backhaul---that do not hold under acoustic constraints.

Research on FL tailored to underwater systems remains limited.
Pei~et~al.~\cite{pei2023fediout} articulated the concept of FL for IoUT and mapped its potential benefits to underwater applications, but provided no concrete algorithm or energy model.
Popli~et~al.~\cite{popli2025fl_underwater_drones} applied FL for intrusion detection in underwater drone networks, focusing on security rather than energy-aware multi-tier aggregation.
Shaheen~et~al.~\cite{shaheen2024fl_iout} surveyed FL's potential for IoUT but did not address cooperative aggregation or reachability-aware communication design.
No existing work provides an end-to-end framework that jointly targets anomaly detection, hierarchical cooperative aggregation, and reachability-aware control under physically grounded underwater acoustic models.

% ------------------------------------------------------------------
\subsection{Communication- and Energy-Efficient Federated Optimisation}
\label{subsec:rw_efficient_fl}

Reducing FL communication cost is critical for bandwidth-scarce acoustic channels.
Gradient sparsification and quantisation~\cite{lin2018dgc}, local-update schemes that increase computation per round~\cite{mcmahan2017communication}, and server-side adaptive optimisers~\cite{reddi2021adaptive_fl} all reduce synchronisation frequency.
In wireless FL, energy-aware scheduling and resource allocation have been studied to minimise battery drain and latency, jointly modelling transmit power, bandwidth, and local computation~\cite{yang2020energy,chen2021energy_fl_convergence}.
Federated anomaly detection studies in IoT have adopted autoencoders and semi-supervised models within FL pipelines~\cite{liu2022intrusion_maritime,aouedi2022fed_semisup_iiot}, but none address the unique energy--delay profile of underwater acoustics.

Adaptive orchestration remains relevant background.
RL has been applied to underwater sensor network routing and to FL client selection or aggregation in heterogeneous wireless networks~\cite{hu2020rl_uwsn_routing,jiang2023drl_uwsn_routing,wang2020fedrl_noniid,nguyen2021fliotsurvey,he2024fedaa_rl_agg}. However, these studies primarily target packet delivery, wireless scheduling, or learned coordination itself rather than the participation-aware IoUT design questions addressed here. We therefore retain RL only as related-work context and focus the main paper on deterministic association and cooperation rules that are easier to interpret, reproduce, and deploy.

% ------------------------------------------------------------------
\subsection{Summary of Gaps and Positioning}
\label{subsec:rw_gaps}

Table~\ref{tab:rw_comparison} summarises representative works and clarifies the positioning of this paper.
Three specific gaps emerge:
\begin{enumerate}
  \item \textbf{Participation-aware comparison is missing:}
    Existing underwater FL studies do not separate \emph{energy efficiency} from \emph{network participation}. In IoUT, this omission is critical because flat FL may exclude sensors that cannot directly reach the gateway.
  \item \textbf{Selective intermediate aggregation is under-studied:}
    Prior HFL work typically assumes either no cooperation or fixed edge cooperation over reliable backhaul. IoUT requires a more careful question: when does fog-to-fog exchange improve learning enough to justify its acoustic cost?
  \item \textbf{Physics-grounded design rules are lacking:}
    Federated anomaly detection exists for terrestrial IoT~\cite{liu2022intrusion_maritime,aouedi2022fed_semisup_iiot}, but no prior work provides IoUT-specific rules for choosing between flat FL, hierarchy, compression, and selective cooperation under realistic acoustic reachability constraints.
\end{enumerate}

% ===== COMPARISON TABLE =====
\begin{table*}[t]
\centering
\small
\caption{Representative related works and positioning of this paper.
(\checkmark: explicitly addressed; $\sim$: partially/implicitly addressed; blank: not addressed.)}
\label{tab:rw_comparison}
\begin{tabularx}{\textwidth}{@{}l X c c c c@{}}
\toprule
\textbf{Reference} & \textbf{Main contribution} & \textbf{IoUT} & \textbf{Anom.~det.} & \textbf{HFL} & \textbf{Energy} \\
\midrule
McMahan et~al.~\cite{mcmahan2017communication} & FedAvg: communication-efficient distributed learning. & & & & $\sim$ \\
Li et~al.~\cite{li2020fedprox} & FedProx: proximal term for heterogeneous FL. & & & & \\
Karimireddy et~al.~\cite{karimireddy2020scaffold} & SCAFFOLD: variance reduction for client drift. & & & & \\
Liu et~al.~\cite{liu2020client} & Client-edge-cloud HFL in cellular networks. & & & \checkmark & $\sim$ \\
Luo et~al.~\cite{luo2020hfel_twc} & HFEL: cost-efficient HFL with edge association. & & & \checkmark & \checkmark \\
Wang et~al.~\cite{wang2022feduc} & Unified clustering for HFL with peer-to-peer exchange. & & & \checkmark & $\sim$ \\
Ghosh et~al.~\cite{ghosh2020ifca} & Clustered FL for non-IID data. & & & $\sim$ & \\
Lin et~al.~\cite{lin2018dgc} & Deep Gradient Compression for distributed training. & & & & $\sim$ \\
Yang et~al.~\cite{yang2020energy} & Energy-efficient FL over wireless networks. & & & & \checkmark \\
Omeke et~al.~\cite{omeke2021dekcs} & Dynamic clustering for energy-efficient underwater sensor networking. & \checkmark & & & \checkmark \\
Pei et~al.~\cite{pei2023fediout} & Conceptual discussion of FL for underwater IoT. & \checkmark & $\sim$ & $\sim$ & $\sim$ \\
Popli et~al.~\cite{popli2025fl_underwater_drones} & FL for intrusion detection in underwater drones. & \checkmark & \checkmark & & $\sim$ \\
Shaheen et~al.~\cite{shaheen2024fl_iout} & Survey on FL potential for IoUT. & \checkmark & & $\sim$ & $\sim$ \\
Liu et~al.~\cite{liu2022intrusion_maritime} & Federated intrusion detection for maritime transport. & $\sim$ & \checkmark & & \\
Aouedi et~al.~\cite{aouedi2022fed_semisup_iiot} & Federated semi-supervised attack detection for industrial IoT. & & \checkmark & & \\
\midrule
\textbf{This work} & \textbf{Participation-aware HFL evaluation with compressed uplinks and selective fog cooperation for IoUT anomaly detection.} & \checkmark & \checkmark & \checkmark & \checkmark \\
\bottomrule
\end{tabularx}
\vspace{-2mm}
\end{table*}

\section{System Model}
\label{sec:system_model}

This section specifies the network architecture, underwater acoustic (UWA) propagation and energy models, and the sensing/data model that underpin the proposed hierarchical learning framework.
All models are simulation-ready and parameterised in Table~\ref{tab:system_params}.

% ------------------------------------------------------------------
\subsection{Network Architecture}
\label{subsec:network_arch}

We consider a 3D underwater deployment volume $\mathcal{V} = [0,L_x]\times[0,L_y]\times[0,H]$, where $z{=}0$ is the sea surface and $z{=}H$ is the maximum depth.
The network comprises three strata (Fig.~\ref{fig:system_setup}):

\begin{itemize}
  \item \textbf{Sensor layer (deep):}
    $N$ stationary sensors $\mathcal{S}=\{s_1,\ldots,s_N\}$ at depths $z\in[z_s^{\min},z_s^{\max}]$.
    Sensors collect multivariate environmental/equipment data and train local anomaly-detection models.
  \item \textbf{Fog layer (mid-water):}
    $M$ fog aggregators $\mathcal{F}=\{f_1,\ldots,f_M\}$ (e.g., AUVs or anchored relays) at depths $z\in[z_f^{\min},z_f^{\max}]$.
    Fog nodes receive and aggregate local model updates and may cooperatively exchange partial aggregates with neighbouring fog nodes.
  \item \textbf{Surface layer:}
    A single surface gateway $g$ at $z{=}0$ hosts the global FL coordinator and maintains a backhaul link to an onshore server.
\end{itemize}

\noindent
Each node $u\in\mathcal{S}\cup\mathcal{F}\cup\{g\}$ has position $\mathbf{p}_u{=}[x_u,y_u,z_u]^\top$.
Sensors and fog nodes are deployed uniformly at random in $(x,y)$ within $\mathcal{V}$ and uniformly in depth within their stratum.
Sensors are static; fog nodes are modelled as quasi-static within a federated round and may drift between rounds according to a Gauss--Markov mobility model with configurable speed.

At each FL round~$t$, sensor $s_i$ associates to a fog node $f_m$ based on link feasibility.
Each modem is subject to a capped source level $\mathrm{SL}_{\max}$, so a directed edge $(u,v)$ exists only if the achievable signal-to-noise ratio (SNR) under that cap satisfies $\mathrm{SNR}_{uv}^{t}(\mathrm{SL}_{\max}) \ge \gamma_{\mathrm{tgt}}$, where $\gamma_{\mathrm{tgt}}$ is the target operating SNR, forming the time-varying communication graph~$\mathcal{G}^{t}$.

% ------------------------------------------------------------------
\subsection{Underwater Acoustic Propagation}
\label{subsec:propagation}

For a link $(u,v)$ of Euclidean distance $d_{uv}=\|\mathbf{p}_u{-}\mathbf{p}_v\|_2$ (m), we adopt a standard large-scale UWA transmission-loss model combining geometric spreading and Thorp absorption~\cite{urick1983principles,thorp1967analytical}:
\begin{equation}
\mathrm{TL}(d,f) = 10\,k\,\log_{10}(d) + \alpha(f)\,\frac{d}{1000},
\label{eq:tl}
\end{equation}
where $k\in[1,2]$ is the spreading factor ($k{=}1.5$ for practical spreading) and $\alpha(f)$ is the Thorp absorption coefficient (dB/km) at carrier frequency $f$~(kHz):
\begin{equation}
\alpha(f) = \frac{0.11\,f^2}{1+f^2} + \frac{44\,f^2}{4100+f^2} + 2.75{\times}10^{-4}f^2 + 0.003.
\label{eq:thorp}
\end{equation}

The acoustic propagation delay is $\tau_{uv} = d_{uv}/c_s$, where $c_s \approx 1500$~m/s is the sound speed.

% ------------------------------------------------------------------
\subsection{Ambient Noise and Receiver SNR}
\label{subsec:noise_snr}

We adopt a Wenz-type ambient noise power spectral density (PSD) model (dB re~$1\;\mu\mathrm{Pa}^2/\mathrm{Hz}$)~\cite{wenz1962acoustic,stojanovic2007underwater} comprising turbulence, shipping ($s\in[0,1]$), wind ($w$~m/s), and thermal components.
The total noise PSD is computed in linear scale and converted back:
\begin{equation}
N_0(f) = 10\log_{10}\!\Bigl(\textstyle\sum_{x}\,10^{N_x(f)/10}\Bigr),
\label{eq:noise_total}
\end{equation}
where $x\in\{\text{turb, ship, wind, therm}\}$ and the per-component models follow the standard Wenz expressions (see~\cite{stojanovic2007underwater} for formulae).
For receiver bandwidth $B$~(Hz), the resulting noise level is $\mathrm{NL}(f,B) = N_0(f) + 10\log_{10}(B)$.

The receiver SNR follows a passive-sonar-equation form:
\begin{equation}
\mathrm{SNR}_{uv} = \mathrm{SL}_u - \mathrm{TL}(d_{uv},f) - \mathrm{NL}(f,B) - \mathrm{IL},
\label{eq:sonar_snr}
\end{equation}
where $\mathrm{SL}_u$ is the source level (dB re $1\;\mu$Pa at 1~m) and $\mathrm{IL}$ is an implementation-loss term.
Omnidirectional transducers are assumed, so the directivity index (DI) is set to zero.

% ------------------------------------------------------------------
\subsection{SNR-Driven Energy Model}
\label{subsec:energy_model}

Energy is the dominant cost in IoUT and must be modelled as a function of link distance and required SNR.
For a feasible link $(u,v)$, the transmitter is assumed to power-control to a target operating SNR $\gamma_{\mathrm{tgt}}$.
The minimum source level required to do so is
\begin{equation}
\mathrm{SL}^{\min}_{u}(u,v) = \gamma_{\mathrm{tgt}} + \mathrm{TL}(d_{uv},f) + \mathrm{NL}(f,B) + \mathrm{IL}.
\label{eq:slmin}
\end{equation}
The capped-source-level feasibility condition is therefore
\begin{equation}
\mathrm{SL}^{\min}_{u}(u,v) \le \mathrm{SL}_{\max}.
\label{eq:sl_feasible}
\end{equation}
The corresponding acoustic transmit power (W) is~\cite{urick1983principles}
\begin{equation}
P_{\mathrm{ac}} = \frac{4\pi\,p_{\mathrm{ref}}^2}{\rho_{\mathrm{w}}\,c_s}\,10^{\mathrm{SL}^{\min}/10},
\label{eq:pac}
\end{equation}
where $\mathrm{SL}^{\min}$ denotes the link-specific minimum source level from Eq.~\eqref{eq:slmin}, $p_{\mathrm{ref}}{=}10^{-6}$~Pa, and $\rho_{\mathrm{w}}\approx1025$~kg/m$^3$ is the water density.
The electrical transmit power is $P_{\mathrm{tx}} = P_{\mathrm{ac}}/\eta_{\mathrm{ea}}$, with $\eta_{\mathrm{ea}}\in(0,1]$ the electro-acoustic efficiency.

For feasible links, we use a Shannon-type rate model $R_{uv} = B\,\log_2\!\bigl(1+10^{\gamma_{\mathrm{tgt}}/10}\bigr)$~(bit/s), which matches the simulator's target-SNR power-control assumption after feasibility has been checked through Eq.~\eqref{eq:sl_feasible}.
The energy to transmit $L$ bits over $(u,v)$ is
\begin{equation}
E_{\mathrm{tx}}(L;u,v) = (P_{\mathrm{tx}} + P_{\mathrm{c,tx}})\,\frac{L}{R_{uv}},
\label{eq:etx}
\end{equation}
where $P_{\mathrm{c,tx}}$ captures circuit overhead.
Receive energy is $E_{\mathrm{rx}} = P_{\mathrm{c,rx}}\cdot L/R_{uv}$.
Computation energy for local training is modelled as $E_{\mathrm{comp},i} = \epsilon_{\mathrm{op}}\,\Phi_i$, where $\Phi_i$ is the number of floating-point operations (FLOPs) and $\epsilon_{\mathrm{op}}$ is the energy per operation~\cite{yang2020energy}.

% ------------------------------------------------------------------
\subsection{Sensing and Data Model}
\label{subsec:data_model}

Each sensor collects multivariate measurements (e.g., temperature, pressure, salinity, vibration), forming a time series mapped to fixed-length feature vectors $\mathbf{x}_{i,n}\in\mathbb{R}^D$.
The local dataset at sensor $s_i$ is $\mathcal{D}_i=\{\mathbf{x}_{i,n}\}_{n=1}^{n_i}$.
We target \emph{unsupervised} anomaly detection using an autoencoder $h_{\bm{\theta}}$ that minimises reconstruction error over normal data:
\begin{equation}
\mathcal{L}_i(\bm{\theta}) = \frac{1}{n_i}\sum_{n=1}^{n_i}\bigl\|\mathbf{x}_{i,n} - h_{\bm{\theta}}(\mathbf{x}_{i,n})\bigr\|_2^2.
\label{eq:ae_loss}
\end{equation}
At inference, samples exceeding a threshold on reconstruction error are flagged as anomalous.

% ------------------------------------------------------------------
\subsection{System Setup Illustration and Parameters}
\label{subsec:system_illustration}

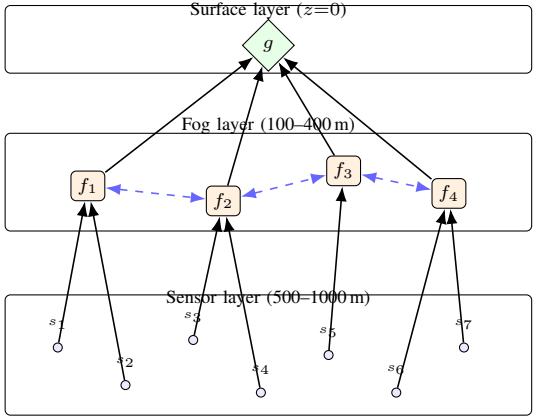
\begin{figure}[t]
\centering
\begin{tikzpicture}[
    node distance=0.9cm, >=Latex,
    sens/.style={circle, draw, fill=blue!8, inner sep=1.2pt},
    fog/.style={rectangle, draw, fill=orange!12, rounded corners=2pt, inner sep=2.5pt},
    surf/.style={diamond, draw, fill=green!10, inner sep=2.8pt},
    layer/.style={draw, rounded corners=2pt, inner sep=4pt, fill=white},
    link/.style={->, line width=0.6pt},
    coop/.style={<->, dashed, line width=0.6pt, blue!60}
]
% --- Surface layer ---
\node[layer, minimum width=7cm, minimum height=0.9cm] (surface) at (0,0) {};
\node[font=\scriptsize] at (0,0.38) {Surface layer ($z{=}0$)};
\node[surf] (gw) at (0,-0.08) {\scriptsize $g$};

% --- Fog layer ---
\node[layer, minimum width=7cm, minimum height=1.3cm] (foglayer) at (0,-1.9) {};
\node[font=\scriptsize] at (0,-1.15) {Fog layer (100--400\,m)};
\node[fog] (f1) at (-2.4,-1.95) {\scriptsize $f_1$};
\node[fog] (f2) at (-0.6,-2.15) {\scriptsize $f_2$};
\node[fog] (f3) at (1.0,-1.75) {\scriptsize $f_3$};
\node[fog] (f4) at (2.4,-2.05) {\scriptsize $f_4$};

% --- Sensor layer ---
\node[layer, minimum width=7cm, minimum height=1.6cm] (senslayer) at (0,-4.2) {};
\node[font=\scriptsize] at (0,-3.45) {Sensor layer (500--1000\,m)};
\foreach \i/\x/\y in {1/-2.8/-4.1, 2/-1.9/-4.6, 3/-1.0/-4.0,
                        4/-0.1/-4.7, 5/0.8/-4.2, 6/1.7/-4.7, 7/2.6/-4.1} {
  \node[sens] (s\i) at (\x,\y) {};
  \node[font=\tiny, above=0.5mm of s\i] {$s_{\i}$};
}

% --- Links ---
\draw[link] (s1) -- (f1); \draw[link] (s2) -- (f1);
\draw[link] (s3) -- (f2); \draw[link] (s4) -- (f2);
\draw[link] (s5) -- (f3);
\draw[link] (s6) -- (f4); \draw[link] (s7) -- (f4);
\draw[link] (f1) -- (gw); \draw[link] (f2) -- (gw);
\draw[link] (f3) -- (gw); \draw[link] (f4) -- (gw);
\draw[coop] (f1) -- (f2); \draw[coop] (f2) -- (f3); \draw[coop] (f3) -- (f4);
\end{tikzpicture}
\vspace{-1mm}
\caption{Stratified IoUT system model.
Sensors (deep layer) transmit local model updates to fog aggregators (mid-water);
fog nodes cooperatively exchange partial aggregates (dashed) and forward to the surface gateway.}
\label{fig:system_setup}
\vspace{-3mm}
\end{figure}

Fig.~\ref{fig:system_setup} illustrates the stratified IoUT topology and HFL communication pattern.
Table~\ref{tab:system_params} lists all baseline parameters; these are used consistently throughout the problem formulation and simulation sections.

\begin{table}[t]
\caption{Baseline system-model and learning parameters; experiment-specific overrides are stated in Section~\ref{sec:evaluation}.}
\label{tab:system_params}
\centering
\small
\begin{tabular}{@{}l@{\hspace{6pt}}l@{\hspace{6pt}}l@{}}
\toprule
\textbf{Parameter} & \textbf{Meaning} & \textbf{Baseline} \\
\midrule
$L_x,\,L_y$ & Horizontal area & $2000\times2000$\,m \\
$H$ & Max depth & 1000\,m \\
$N$ & Sensor nodes & 100 (varied: 50--200) \\
$M$ & Fog aggregators & 10 (varied: 5--20) \\
$z_s^{\min},\,z_s^{\max}$ & Sensor depth range & 500--1000\,m \\
$z_f^{\min},\,z_f^{\max}$ & Fog depth range & 100--400\,m \\
$f$ & Carrier frequency & 12\,kHz \\
$B$ & Receiver bandwidth & 4\,kHz \\
$k$ & Spreading factor & 1.5 \\
$c_s$ & Sound speed & 1500\,m/s \\
$w,\,s$ & Wind speed / shipping & 5\,m/s, 0.5 \\
$\gamma_{\mathrm{tgt}}$ & Target SNR & 10\,dB \\
$\mathrm{IL}$ & Implementation loss & 2\,dB \\
$SL_{\max}$ & Source-level cap & 140\,dB re 1$\mu$Pa @ 1\,m \\
$\eta_{\mathrm{ea}}$ & Electro-acoustic eff. & 0.25 \\
$P_{\mathrm{c,tx}},\,P_{\mathrm{c,rx}}$ & Circuit power & 50\,mW, 30\,mW \\
$E_{\mathrm{init}}$ & Initial battery / sensor & 500\,J \\
$D$ & Feature dimension & 32 \\
AE structure & Hidden layers & [32, 16, 8, 16, 32] \\
$d$ & Model parameters & ${\approx}1350$ \\
$E$ & Local epochs / round & 5 \\
$T$ & FL rounds (exp.-specific) & 20 synth., 30 real \\
$\eta$ & Learning rate & 0.01 \\
\bottomrule
\end{tabular}
\vspace{-2mm}
\end{table}

\section{Problem Formulation}
\label{sec:problem_formulation}

Building on the system model in Section~\ref{sec:system_model}, we formalise the joint anomaly-detection learning task and network-control problem.
The key challenge is to \emph{simultaneously} learn an accurate anomaly detector from distributed, privacy-sensitive sensory data while \emph{minimising} communication energy and delay over time-varying UWA links.

% ------------------------------------------------------------------
\subsection{Learning Objective}
\label{subsec:learning_objective}

Each sensor $i\!\in\!\mathcal{S}$ holds a private dataset $\mathcal{D}_i$ and trains an autoencoder with parameters $\bm{\theta}\in\mathbb{R}^{d}$ to minimise reconstruction loss (Eq.~\eqref{eq:ae_loss}).
The local objective is
\begin{equation}
F_i(\bm{\theta}) = \frac{1}{n_i}\sum_{n=1}^{n_i}\bigl\|\mathbf{x}_{i,n} - h_{\bm{\theta}}(\mathbf{x}_{i,n})\bigr\|_2^2,
\label{eq:local_obj}
\end{equation}
and the population-weighted global objective is
\begin{equation}
F(\bm{\theta}) = \sum_{i\in\mathcal{S}} \frac{n_i}{\sum_{k\in\mathcal{S}} n_k}\, F_i(\bm{\theta}).
\label{eq:global_obj}
\end{equation}
At inference, a sample is flagged anomalous if its reconstruction error exceeds a threshold $\tau_A$ calibrated on a normal-only validation window.

% ------------------------------------------------------------------
\subsection{Hierarchical FL with Cooperative Fog Aggregation}
\label{subsec:hfl_formulation}

Training proceeds in $T$ federated rounds.
At round~$t$, each sensor $i$ is associated to exactly one fog node via decision $a_i^{t}\in\mathcal{F}$, inducing clusters $\mathcal{C}_m^{t} = \{i\in\mathcal{S}: a_i^{t}=m\}$.
The HFL workflow has three stages:

\subsubsection{Local training and compression}
Sensor $i$ performs $E$ local stochastic gradient descent (SGD) epochs starting from the current global model $\bm{\theta}^{t}$:
\begin{equation}
\bm{\theta}_i^{t+1} \leftarrow \bm{\theta}^{t} - \eta\,\nabla F_i(\bm{\theta}^{t};\mathcal{B}_i^{t}),
\label{eq:local_sgd}
\end{equation}
where $\eta$ is the learning rate and $\mathcal{B}_i^{t}$ denotes mini-batches from $\mathcal{D}_i$.
The update $\Delta\bm{\theta}_i^{t} = \bm{\theta}_i^{t+1} - \bm{\theta}^{t}$ is compressed with ratio $\rho\in(0,1]$ (sparsification/quantisation~\cite{lin2018dgc}), yielding a payload of $L_u = \rho\,b\,d$ bits ($b{=}32$ for full precision).

\subsubsection{Fog-level aggregation}
Fog node $m$ aggregates updates from its cluster:
\begin{equation}
\bm{\theta}_m^{t+\tfrac{1}{2}} = \bm{\theta}^{t} + \sum_{i\in\mathcal{C}_m^{t}} \frac{n_i}{\sum_{k\in\mathcal{C}_m^{t}} n_k}\,\Delta\bm{\theta}_i^{t}.
\label{eq:fog_agg}
\end{equation}

\subsubsection{Cooperative fog mixing}
To mitigate non-IID-induced drift and improve robustness to fog drop-out, each fog node $m$ selects up to $K$ neighbours $\mathcal{N}_m^{t}\subseteq\mathcal{F}\setminus\{m\}$ and mixing coefficients $\bm{\alpha}_m^{t}$ satisfying
\begin{equation}
\alpha_{m,j}^{t}\ge 0,\quad \sum_{j\in\{m\}\cup\mathcal{N}_m^{t}} \alpha_{m,j}^{t}=1,\quad |\mathcal{N}_m^{t}|\le K.
\label{eq:mixing}
\end{equation}
The cooperatively mixed model is
\begin{equation}
\tilde{\bm{\theta}}_m^{t+1} = \sum_{j\in\{m\}\cup\mathcal{N}_m^{t}} \alpha_{m,j}^{t}\,\bm{\theta}_j^{t+\tfrac{1}{2}}.
\label{eq:coop_mix}
\end{equation}

\subsubsection{Surface-level global aggregation}
The surface gateway fuses the cooperatively aggregated fog models:
\begin{equation}
\bm{\theta}^{t+1} = \sum_{m\in\mathcal{F}} \frac{\sum_{i\in\mathcal{C}_m^{t}} n_i}{\sum_{k\in\mathcal{S}} n_k}\,\tilde{\bm{\theta}}_m^{t+1}.
\label{eq:global_agg}
\end{equation}

% ------------------------------------------------------------------
\subsection{Per-Round Energy and Latency Cost}
\label{subsec:cost_model}

Using the SNR-driven energy model (Eqs.~\eqref{eq:slmin}--\eqref{eq:etx}), the per-round communication energy decomposes into three components:

\noindent
\emph{Sensor-to-fog upload:}
\begin{equation}
E_{\mathrm{s2f}}^{t} = \sum_{i\in\mathcal{S}} E_{\mathrm{tx}}(L_u;\,i,a_i^t).
\label{eq:e_s2f}
\end{equation}

\noindent
\emph{Fog-to-fog cooperation:}
\begin{equation}
E_{\mathrm{f2f}}^{t} = \sum_{m\in\mathcal{F}}\sum_{j\in\mathcal{N}_m^{t}} E_{\mathrm{tx}}(L_f;\,m,j),
\label{eq:e_f2f}
\end{equation}
where $L_f$ is the fog-exchange payload size.

\noindent
\emph{Fog-to-surface uplink:}
\begin{equation}
E_{\mathrm{f2g}}^{t} = \sum_{m\in\mathcal{F}} E_{\mathrm{tx}}(L_g;\,m,g).
\label{eq:e_f2g}
\end{equation}
Here $L_g$ denotes the fog-to-gateway payload size.

The total round energy is
\begin{equation}
E_{\mathrm{round}}^{t} = E_{\mathrm{s2f}}^{t} + E_{\mathrm{f2f}}^{t} + E_{\mathrm{f2g}}^{t},
\label{eq:e_round}
\end{equation}
and the round latency (dominated by the slowest parallel link plus propagation delay) is
\begin{equation}
\tau_{\mathrm{round}}^{t} = \max\Bigl\{\max_{i}\tau_{i\to a_i^t}^{t},\;\max_{m,j}\tau_{m\to j}^{t},\;\max_{m}\tau_{m\to g}^{t}\Bigr\} + \tau_{\mathrm{comp}}^{t},
\label{eq:tau_round}
\end{equation}
where $\tau_{u\to v}^{t}=d_{uv}^{t}/c_s + L_{u\to v}^{t}/R_{uv}^{t}$ denotes the latency of an individual link transmission, $L_{u\to v}^{t}$ is the corresponding payload on link $(u,v)$, and $\tau_{\mathrm{comp}}^{t}$ is the maximum local-computation time among active participants in round~$t$.
Battery dynamics track energy depletion: if $E_i^{t}$ denotes the residual battery of sensor~$i$ at the start of round~$t$, then $E_i^{t+1} = E_i^{t} - E_{\mathrm{tx}}(L_u;\,i,a_i^t) - E_{\mathrm{comp},i}^{t}$.

% ------------------------------------------------------------------
\subsection{Joint Optimisation Objective}
\label{subsec:optimisation}

We seek to jointly optimise the anomaly-detector parameters $\bm{\theta}$ and the control decisions (associations $\{a_i^t\}$, cooperation sets $\{\mathcal{N}_m^t, \bm{\alpha}_m^t\}$) that minimise energy and delay while promoting fast convergence:
\begin{align}
\min_{\{\bm{\theta}^{t}\},\,\{a_i^{t}\},\,\{\mathcal{N}_m^{t},\bm{\alpha}_m^{t}\}} \quad
& F(\bm{\theta}^{T}) + \lambda_E \sum_{t=0}^{T-1} E_{\mathrm{round}}^{t} \nonumber\\
&\qquad + \lambda_\tau \sum_{t=0}^{T-1} \tau_{\mathrm{round}}^{t}
\label{eq:main_obj} \\
\text{s.t.}\quad
& a_i^{t}\in\mathcal{F}, \quad \forall\,i,t, \label{eq:c_assoc}\\
& \bm{\alpha}_m^{t}\;\text{satisfies~\eqref{eq:mixing}}, \quad \forall\,m,t, \label{eq:c_alpha}\\
& E_i^{t+1} \ge E_{\min}, \quad \forall\,i,t, \label{eq:c_energy}\\
& \tau_{\mathrm{round}}^{t} \le \tau_{\max}, \quad \forall\,t, \label{eq:c_latency}\\
& \mathrm{SNR}_{\ell}^{t} \ge \gamma_{\mathrm{tgt}}, \quad \forall\;\ell,t. \label{eq:c_snr}
\end{align}
Here $\lambda_E$ and $\lambda_\tau$ are energy and latency weighting coefficients, $E_{\min}$ is the minimum battery reserve, $\tau_{\max}$ is the per-round latency deadline, and $\ell$ indexes an active communication link.

Problem~\eqref{eq:main_obj}--\eqref{eq:c_snr} is non-convex and combinatorial (discrete associations, variable neighbour sets) coupled with non-convex learning dynamics.
To keep the framework transparent and practically deployable, this paper studies a compact family of \emph{interpretable} decision rules over the association decisions $\{a_i^t\}$ and cooperation sets $\{\mathcal{N}_m^t,\bm{\alpha}_m^t\}$.
This highlights the three questions that matter most in IoUT: which sensors can participate at all, whether fog nodes should cooperate, and how much energy that cooperation is allowed to consume.

% ------------------------------------------------------------------
\subsection{Decision Space}
\label{subsec:policy_space}

The decision space considered in this paper is intentionally small and deployment-oriented.
For \emph{flat FL}, only sensors with a feasible direct sensor-to-gateway link can participate.
For \emph{hierarchical FL}, each sensor is associated with its nearest feasible fog. If a sensor temporarily has no feasible fog link, it is inactive for that round; this case does not arise in the reported synthetic settings, where every sensor retains at least one feasible fog path.

We study three fog-level cooperation rules:
\begin{enumerate}
  \item \textbf{HFL-NoCoop:} each fog aggregates only its own cluster and forwards the result to the gateway.
  \item \textbf{HFL-Nearest:} each fog cooperates with its nearest feasible fog neighbour using fixed mixing weights.
  \item \textbf{HFL-Selective:} only smaller fog clusters cooperate, and only with one nearby larger feasible neighbour.
\end{enumerate}

The selective rule is designed to answer a simple IoUT question: if fog-to-fog exchange is expensive, can we restrict it to the cases where it is most likely to reduce cluster imbalance?
Section~\ref{subsec:policy_families} gives the exact rule used in the simulator.
The same framework could support more adaptive learned orchestrators, including RL-based ones, but the main paper focuses on deterministic decision rules because they provide a transparent and reproducible basis for IoUT design analysis.

% ------------------------------------------------------------------
\subsection{Remark on Convergence}
\label{subsec:convergence_remark}

The cooperative fog mixing in~\eqref{eq:coop_mix} can be viewed as a gossip-style averaging step superimposed on standard hierarchical FedAvg.
Under mild assumptions (strongly convex or Polyak--Lojasiewicz (PL) local objectives, bounded gradient variance, and a doubly stochastic mixing matrix induced by $\bm{\alpha}_m^{t}$), hierarchical FL with periodic consensus has been shown to converge at rate $O(1/T)$ with a bias controlled by the local-update count $E$ and the spectral gap of the mixing topology~\cite{lian2017decentralized_sgd,wang2022feduc,luo2020hfel_twc}.
Our selective rule preserves this structure while restricting the communication graph to a sparse, feasible subset of fog-to-fog links.
Accordingly, the main question in this paper is not whether convergence is possible in principle, but how different feasible rule sets trade off convergence quality, participation coverage, and acoustic energy.

% ---------- Bi-level overview figure ----------
\begin{figure}[t]
\centering
% Resizebox ensures the figure never exceeds the column width
\resizebox{\columnwidth}{!}{
\begin{tikzpicture}[
    font=\scriptsize, 
    node distance=6mm and 4mm, % Reduced horizontal spacing to 4mm
    box/.style={draw, rounded corners, align=center, inner sep=3pt,
                minimum width=16mm, minimum height=8mm}, % Slightly narrower min-width
    arr/.style={-Latex, thick}
]

% ----- Nodes -----
\node[box] (local) {Local training\\(sensors)};
\node[box, right=of local] (upload) {Compressed\\upload $L_u$};
\node[box, right=of upload] (fog) {Fog agg.\\Eq.~\eqref{eq:fog_agg}};
\node[box, right=of fog] (coop) {Coop.\ mixing\\Eq.~\eqref{eq:coop_mix}};
\node[box, right=of coop] (surf) {Global agg.\\Eq.~\eqref{eq:global_agg}};

% Decision-rule block positioned relative to the center of the flow
\node[box, below=8mm of coop, fill=gray!8, draw=gray!55, minimum width=31mm] (rules)
  {Association \&\\cooperation rules\\$a_i^{t}$, $(\mathcal{N}_m^{t},\bm{\alpha}_m^{t})$};

% ----- Edges -----
\draw[arr] (local) -- (upload);
\draw[arr] (upload) -- (fog);
\draw[arr] (fog) -- (coop);
\draw[arr] (coop) -- (surf);

% Control loops
\draw[arr, gray!60, dashed] (rules.west) -| (upload.south);
\draw[arr, gray!60, dashed] (rules) -- (coop.south);

\end{tikzpicture}
} % End of resizebox
\vspace{-1mm}
\caption{Bi-level view of the framework: hierarchical FL training (top) and the decision-rule layer that selects sensor associations and fog cooperation patterns (bottom).}
\label{fig:problem_overview}
\vspace{-3mm}
\end{figure}
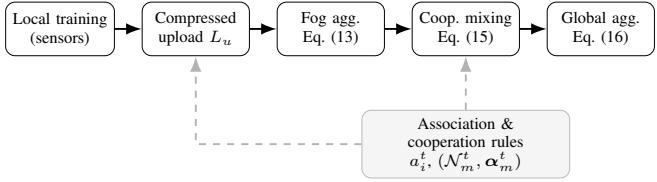

% Section V: Proposed framework
\section{Proposed Hierarchical Federated Learning Framework}
\label{sec:framework}
Building on the system model and optimisation problem defined in the previous sections, this section presents the proposed hierarchical federated learning method used throughout the paper. We describe the end-to-end training pipeline (Section~\ref{subsec:pipeline}), the association and cooperation rules that implement selective aggregation (Section~\ref{subsec:policy_families}), the update-compression strategy that keeps acoustic communication tractable (Section~\ref{subsec:compression}), and the post-convergence anomaly-threshold calibration procedure (Section~\ref{subsec:threshold}).

\subsection{Overview of the Training Pipeline}
\label{subsec:pipeline}
Algorithm~\ref{alg:hfl_rca} summarises one federated round of the framework.
At the start of each round~$t$, the selected association rule assigns a fog node $a_i^{t}$ to each sensor and the selected cooperation rule determines the fog neighbour sets $\mathcal{N}_m^{t}$ with mixing weights $\bm{\alpha}_m^{t}$.
Sensors then perform $E$ local SGD epochs on the current global model $\bm{\theta}^{t}$ and compress the resulting update $\Delta\bm{\theta}_i^{t}$ via the Top-$K$ sparsification--quantisation pipeline described in Section~\ref{subsec:compression}.
The compressed payload is transmitted acoustically to the assigned fog node.
Each fog node aggregates its cluster's updates according to Eq.~\eqref{eq:fog_agg}, optionally mixes with selected neighbours via Eq.~\eqref{eq:coop_mix}, and uploads the result to the surface gateway.
The gateway performs weighted global aggregation (Eq.~\eqref{eq:global_agg}) and broadcasts the new model $\bm{\theta}^{t+1}$ back through the hierarchy.
Per-round energy, latency, and participation are then recorded for evaluation.

\begin{algorithm}[t]
\caption{Proposed HFL Main Loop (per FL round $t$)}
\label{alg:hfl_rca}
\small
\begin{algorithmic}[1]
\REQUIRE Current global model $\bm{\theta}^{t}$; association rule $\mathcal{R}_{\mathrm{assoc}}$; cooperation rule $\mathcal{R}_{\mathrm{coop}}$
\ENSURE Updated global model $\bm{\theta}^{t+1}$
\STATE \textbf{Association and cooperation decisions:}
\FOR{each sensor $i \in \mathcal{S}$ \textbf{in parallel}}
  \STATE $a_i^{t} \leftarrow \mathcal{R}_{\mathrm{assoc}}(i)$ \COMMENT{fog association}
\ENDFOR
\FOR{each fog node $m \in \mathcal{F}$ \textbf{in parallel}}
  \STATE $(\mathcal{N}_m^{t},\,\bm{\alpha}_m^{t}) \leftarrow \mathcal{R}_{\mathrm{coop}}(m,\{\mathcal{C}_j^{t}\})$
\ENDFOR
\STATE \textbf{Local training \& compression:}
\FOR{each sensor $i \in \mathcal{S}$ \textbf{in parallel}}
  \STATE $\bm{\theta}_i^{t+1} \leftarrow \textsc{LocalSGD}(\bm{\theta}^{t},\,\mathcal{D}_i,\,E,\,\eta)$
  \STATE $\widetilde{\Delta\bm{\theta}}_i^{t} \leftarrow \textsc{Compress}(\Delta\bm{\theta}_i^{t},\,\bm{e}_i^{t},\,\rho)$ \COMMENT{Sec.~\ref{subsec:compression}}
  \STATE Transmit $\widetilde{\Delta\bm{\theta}}_i^{t}$ to fog $a_i^{t}$
\ENDFOR
\STATE \textbf{Fog aggregation \& cooperation:}
\FOR{each fog $m \in \mathcal{F}$ \textbf{in parallel}}
  \STATE $\bm{\theta}_m^{t+\frac{1}{2}} \leftarrow$ Eq.~\eqref{eq:fog_agg} \COMMENT{intra-cluster aggregation}
  \STATE $\tilde{\bm{\theta}}_m^{t+1} \leftarrow$ Eq.~\eqref{eq:coop_mix} using $(\mathcal{N}_m^{t},\,\bm{\alpha}_m^{t})$
\ENDFOR
\STATE \textbf{Global aggregation:}
\STATE $\bm{\theta}^{t+1} \leftarrow$ Eq.~\eqref{eq:global_agg}
\STATE Broadcast $\bm{\theta}^{t+1}$ to all fog nodes and sensors
\STATE Record round energy, latency, and participation statistics
\end{algorithmic}
\end{algorithm}

\subsection{Association and Cooperation Rules}
\label{subsec:policy_families}

The simulator instantiates the following practical rule families.

\textbf{Flat FL baselines.}
FedAvg and FedProx are treated as direct-to-gateway baselines.
Only sensors with a feasible sensor-to-gateway link are allowed to upload updates, so their effective participation set is a topology-dependent subset of $\mathcal{S}$.

\textbf{Nearest-feasible hierarchical association.}
All hierarchical policies use the same association rule: sensor $i$ attaches to its nearest feasible fog node.
This makes the comparison between hierarchical variants depend only on the \emph{inter-fog} cooperation rule, not on a confounded association choice.

\textbf{No cooperation.}
For HFL-NoCoop, each fog simply forwards its intra-cluster aggregate, i.e., $\mathcal{N}_m^{t}=\varnothing$ and $\tilde{\bm{\theta}}_m^{t+1}=\bm{\theta}_m^{t+\frac{1}{2}}$.
This provides the lowest-energy hierarchical baseline and isolates the value of any fog-to-fog exchange.

\textbf{Always-on nearest-neighbour cooperation.}
For HFL-Nearest, each fog cooperates with its nearest feasible fog neighbour using fixed mixing weights $(0.7, 0.3)$.
This is a natural but potentially wasteful baseline because it forces fog-to-fog exchange even when the receiving cluster is already large and statistically diverse.

\textbf{Selective cooperation.}
For HFL-Selective, let $c_m^{t}=|\mathcal{C}_m^{t}|$ denote the cluster size at fog $m$ and let $\bar{c}^{t}$ be the mean of the non-empty cluster sizes.
A fog is eligible to cooperate only if
\begin{equation}
c_m^{t} \le \max\{2,\;0.75\,\bar{c}^{t}\}.
\label{eq:small_cluster_rule}
\end{equation}
Among its feasible neighbours, fog $m$ then considers only those with larger cluster size and with fog-to-fog distance below the first quartile of feasible fog distances.
If such neighbours exist, $m$ selects the nearest one, say $j$, and applies
\begin{equation}
\tilde{\bm{\theta}}_m^{t+1} = 0.8\,\bm{\theta}_m^{t+\frac{1}{2}} + 0.2\,\bm{\theta}_j^{t+\frac{1}{2}}.
\label{eq:selective_mix}
\end{equation}
Otherwise, it falls back to no cooperation.
This rule is intentionally simple: it uses cooperation only to help smaller, less representative clusters borrow information from nearby larger ones.

% ------------------------------------------------------------------
\subsection{Update Compression Strategy}
\label{subsec:compression}

Acoustic bandwidth is the most stringent bottleneck in IoUT; even modest autoencoders ($d{\approx}1350$ parameters) produce multi-kilobit payloads at full precision.
We therefore apply a two-stage compression pipeline to the local model update $\Delta\bm{\theta}_i^{t}$.

\subsubsection{Top-$K$ sparsification with error feedback}
Only the $K{=}\lceil\rho_s\,d\rceil$ coordinates of largest magnitude are retained, where $\rho_s\in(0,1]$ is the sparsification ratio.
The discarded coordinates are accumulated in a local error buffer $\bm{e}_i^{t}$ and added back at the next round, following the error-feedback (EF) mechanism~\cite{stich2018sparsified,alistarh2018sparsified}:
\begin{equation}
\bm{v}_i^{t} = \Delta\bm{\theta}_i^{t} + \bm{e}_i^{t-1},\quad
\widetilde{\bm{v}}_i^{t} = \mathrm{Top\text{-}}K(\bm{v}_i^{t}),\quad
\bm{e}_i^{t} = \bm{v}_i^{t} - \widetilde{\bm{v}}_i^{t},
\label{eq:topk_ef}
\end{equation}
where $\mathrm{Top\text{-}}K(\cdot)$ zeros all but the $K$ largest-magnitude entries.
Error feedback ensures that no gradient information is permanently lost, preserving convergence guarantees under biased compression~\cite{lin2018dgc}.

\subsubsection{Post-sparsification quantisation}
The non-zero entries of $\widetilde{\bm{v}}_i^{t}$ are quantised to 8-bit fixed-point representation~\cite{dettmers2016bit}, further reducing per-entry cost from 32 to 8~bits.
Combined with sparsification, the effective payload becomes
\begin{equation}
L_u = \rho_s\,d\,(b_q + b_{\mathrm{idx}}) \;\;\text{bits},
\label{eq:payload}
\end{equation}
where $b_q{=}8$ is the quantisation bit-width and $b_{\mathrm{idx}}{=}\lceil\log_2 d\rceil$ bits encode each coordinate index.
For the baseline autoencoder ($d{\approx}1350$, $b_{\mathrm{idx}}{=}11$), setting $\rho_s{=}0.05$ yields an effective compression ratio $\rho \approx 0.03$ relative to uncompressed 32-bit transmission, reducing the per-sensor payload from ${\sim}43$\,kbit to ${\sim}1.3$\,kbit per round.

% ------------------------------------------------------------------
\subsection{Anomaly Threshold Calibration}
\label{subsec:threshold}

After the federated model $\bm{\theta}^{*}$ has converged, each sensor (or the global coordinator) calibrates the anomaly decision threshold on a held-out normal-only validation window $\mathcal{D}_i^{\mathrm{val}}$.
The reconstruction error for each validation sample is computed as $e_{i,n} = \|\mathbf{x}_{i,n} - h_{\bm{\theta}^{*}}(\mathbf{x}_{i,n})\|_2^2$, and the threshold is set at the $p$-th percentile of the error distribution:
\begin{equation}
\tau_{A,i} = \mathrm{Percentile}\bigl(\{e_{i,n}\}_{n=1}^{|\mathcal{D}_i^{\mathrm{val}}|},\;p\bigr),
\label{eq:threshold}
\end{equation}
where $p$ controls the trade-off between false-positive rate and detection sensitivity~\cite{pang2021deep_ad_survey}; we use $p{=}99$ in all experiments.
At inference, any sample with $e_{i,n}>\tau_{A,i}$ is flagged as anomalous.

In the \emph{per-sensor} variant, each sensor $i$ maintains its own $\tau_{A,i}$, which naturally adapts to local data distributions and is preferable under strong non-IID conditions.
In the \emph{global} variant, reconstruction errors are pooled across all sensors and a single threshold $\tau_A$ is applied network-wide, simplifying deployment at the cost of reduced per-node specificity.
In this work, we adopt the global variant for simplicity and evaluate it in Section~\ref{sec:evaluation}.

% --- Section VI: Performance Evaluation ---
\section{Performance Evaluation}
\label{sec:evaluation}

This section evaluates the proposed framework from three complementary perspectives. First, we study how network scale and acoustic reachability influence the relative merits of flat and hierarchical learning. Second, we quantify the energy impact of selective cooperative aggregation and compressed uploads. Third, we validate the approach on real anomaly-detection benchmarks to test whether the observed trends persist beyond the synthetic setting.

% ------------------------------------------------------------------
\subsection{Evaluation Setup}
\label{subsec:eval_setup}

We use a custom Python/PyTorch simulator that implements the capped-source-level acoustic model from Section~\ref{sec:system_model}.
The anomaly detector is a lightweight symmetric autoencoder ($32{\to}16{\to}8{\to}16{\to}32$, ${\approx}1{,}352$ parameters).
Unless otherwise stated, sensors upload compressed model updates using Top-$K$ sparsification ($\rho_s{=}0.05$) with 8-bit quantisation, while fog-to-fog and fog-to-gateway transmissions use full-precision exchange.
Thresholds are calibrated from held-out validation data using the 99th percentile rule in Eq.~\eqref{eq:threshold}.

\begin{figure}[t]
\centering
\includegraphics[width=\linewidth]{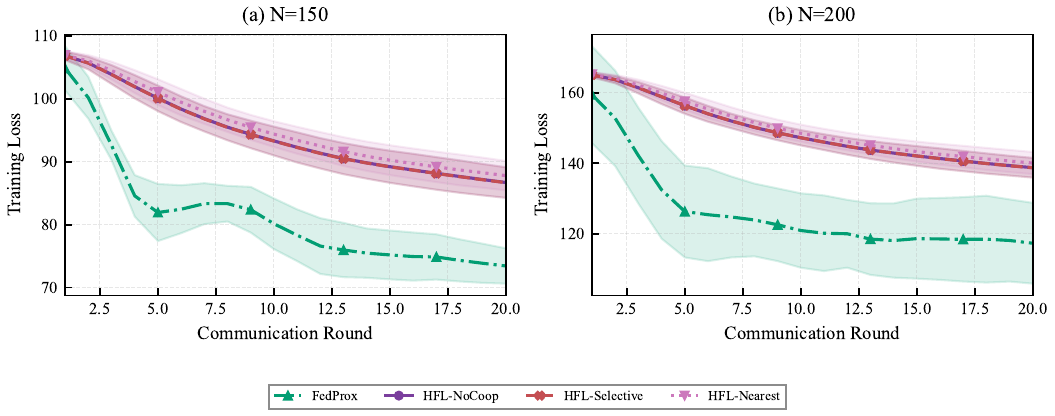}
\vspace{-5mm}
\caption{Convergence behaviour of the synthetic method set used in the main scalability study.
(a) Training loss at $N{=}150$.
(b) Training loss at $N{=}200$.
Across both scales, the loss curves flatten by roughly rounds 10--12, supporting the choice $T{=}20$ for the main synthetic experiments.
Error bands denote one standard deviation over three seeds.}
\label{fig:convergence_check}
\end{figure}

The synthetic study uses three seeds and emphasises the large-scale regime in which effective participation becomes important.
For tractability, the main synthetic experiments use $T{=}20$ rounds with $N\in\{50,150,200\}$ and $M{=}N/10$.
Figure~\ref{fig:convergence_check} shows a dedicated convergence check for the synthetic method set used in the main scalability study; at both $N{=}150$ and $N{=}200$, the loss curves flatten by roughly rounds 10--12, so $T{=}20$ provides a margin beyond the observed plateau.
The real-data study uses $T{=}30$ rounds on SMD, SMAP, and MSL to provide a more conservative budget on the longer benchmark sequences.

\subsection{Baselines}
\label{subsec:baselines}

We focus on the methods that are most informative for IoUT system design:
\begin{enumerate}
  \item \textbf{Centralised:} all-data oracle reference at the gateway; included only as a real-data reference and not as an underwater-feasible deployment option.
  \item \textbf{FedAvg}~\cite{mcmahan2017communication}: direct star-topology FL over feasible sensor-to-gateway links.
  \item \textbf{FedProx}~\cite{li2020fedprox}: FedAvg with a proximal term; this is the strongest flat baseline in our large-scale experiments.
  \item \textbf{HFL-NoCoop:} nearest-feasible-fog association with no fog-to-fog exchange.
  \item \textbf{HFL-Selective:} nearest-feasible-fog association with the selective cooperation rule in Eq.~\eqref{eq:selective_mix}.
  \item \textbf{HFL-Nearest:} nearest-feasible-fog association with always-on neighbour cooperation.
\end{enumerate}

Variance-reduced baselines such as SCAFFOLD remain relevant under heterogeneous data, but exploratory simulations in this problem setting showed unstable behaviour under severe heterogeneity, so the main comparison concentrates on the most robust deployable baselines; the broader traces will be included in the released code and results package.

\subsection{Scalability Under Acoustic Reachability Constraints}
\label{subsec:exp_scalability}

\begin{figure}[t]
\centering
\includegraphics[width=\linewidth]{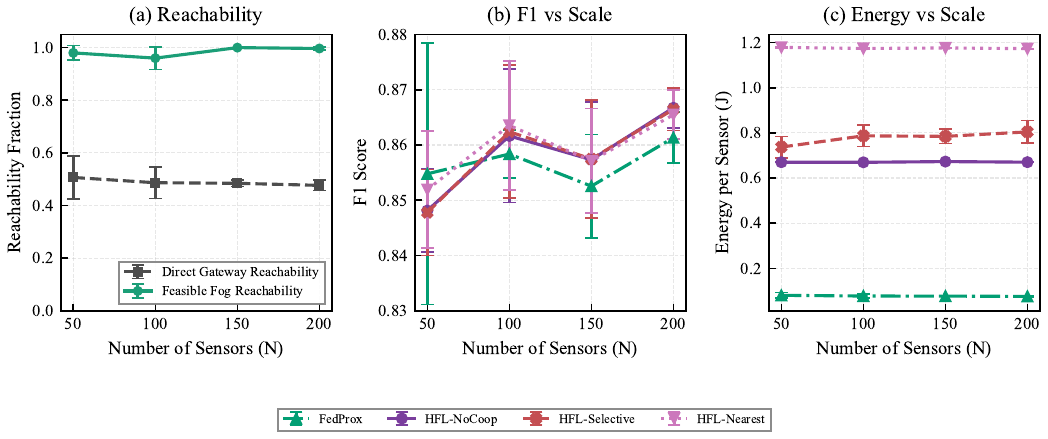}
\vspace{-5mm}
\caption{Participation and trade-off results in synthetic IoUT deployments.
(a) Direct gateway reachability decreases with scale, while fog-assisted reachability remains near-complete.
(b) Detection quality as a function of network scale.
(c) Per-sensor communication energy as a function of network scale.
Error bars denote one standard deviation over three seeds.}
\label{fig:recovery_participation}
\end{figure}

We begin with the scalability question that motivates hierarchical learning in IoUT. Figure~\ref{fig:recovery_participation} shows that the main distinction between flat FL and hierarchical FL is not only communication cost, but also \emph{effective network participation in training}.
At $N{=}50$, only about 51\% of sensors can directly reach the gateway, whereas roughly 98\% have at least one feasible fog.
At $N{=}100$, direct gateway reachability is still only about 49\%, while feasible fog reachability is already about 96\%.
At $N{=}150$ and $N{=}200$, direct gateway reachability stays near 48\%, while feasible fog reachability becomes essentially full.
This means that the flat baselines optimise over a topology-dependent subset of the network, whereas hierarchical methods continue to include almost all sensors.

Table~\ref{tab:scale_tradeoff} compares the strongest flat baseline (FedProx) with the three hierarchical methods across the transition from moderate to large scale.
At $N{=}50$, FedProx remains the cheapest option and is slightly higher in mean F1, which is consistent with the still-modest participation gap between direct and fog-assisted training.
By $N{=}100$, however, hierarchy already recovers about 0.96 participation versus 0.49 for FedProx, and the hierarchical methods become slightly stronger in mean F1.
At $N{=}200$, FedProx still achieves the lowest total energy (15.2\,J) but only 0.48 participation, whereas HFL-NoCoop, HFL-Selective, and HFL-Nearest all preserve full participation and achieve slightly higher mean F1.
Among the hierarchical methods, HFL-NoCoop is the strongest default rule set in this study: it matches or exceeds the other HFL variants in F1 score while avoiding the cost of unnecessary fog-to-fog exchange.
These synthetic F1 gaps are small relative to the three-seed variation; the more robust result is the participation-versus-energy trade-off.

\begin{table}[t]
\caption{Scalability results under acoustic reachability constraints.
Gateway reachability falls from 0.51 at $N{=}50$ to about 0.48 at $N{\geq}150$, while feasible fog reachability rises from 0.98 at $N{=}50$ to essentially full coverage by $N{\geq}150$.
Mean$\pm$std over three seeds.}
\label{tab:scale_tradeoff}
\centering
\footnotesize
\begin{tabular}{@{}ccccc@{}}
\toprule
\textbf{$N$} & \textbf{Method} & \textbf{Participation} & \textbf{F1 score} & \textbf{Energy (J)} \\
\midrule
\multirow{4}{*}{50}
& FedProx        & 0.51 & 0.8548$\pm$0.0236 &   4.0$\pm$0.7 \\
& HFL-NoCoop     & 0.98 & 0.8481$\pm$0.0075 &  33.5$\pm$0.1 \\
& HFL-Selective  & 0.98 & 0.8478$\pm$0.0078 &  36.9$\pm$2.4 \\
& HFL-Nearest    & 0.98 & 0.8519$\pm$0.0106 &  58.9$\pm$0.1 \\
\midrule
\multirow{4}{*}{100}
& FedProx        & 0.49 & 0.8584$\pm$0.0044 &   7.7$\pm$1.0 \\
& HFL-NoCoop     & 0.96 & 0.8616$\pm$0.0121 &  67.0$\pm$0.1 \\
& HFL-Selective  & 0.96 & 0.8624$\pm$0.0120 &  78.7$\pm$4.7 \\
& HFL-Nearest    & 0.96 & 0.8635$\pm$0.0117 & 117.3$\pm$0.1 \\
\midrule
\multirow{4}{*}{150}
& FedProx        & 0.48 & 0.8526$\pm$0.0094 & 11.6$\pm$0.3 \\
& HFL-NoCoop     & 1.00 & 0.8573$\pm$0.0104 & 100.9$\pm$0.2 \\
& HFL-Selective  & 1.00 & 0.8575$\pm$0.0106 & 117.7$\pm$4.7 \\
& HFL-Nearest    & 1.00 & 0.8571$\pm$0.0095 & 176.4$\pm$0.2 \\
\midrule
\multirow{4}{*}{200}
& FedProx        & 0.48 & 0.8613$\pm$0.0046 & 15.2$\pm$0.6 \\
& HFL-NoCoop     & 1.00 & 0.8667$\pm$0.0036 & 134.0$\pm$0.2 \\
& HFL-Selective  & 1.00 & 0.8664$\pm$0.0038 & 160.8$\pm$10.1 \\
& HFL-Nearest    & 1.00 & 0.8655$\pm$0.0044 & 234.6$\pm$0.2 \\
\bottomrule
\end{tabular}
\vspace{-2mm}
\end{table}

Two findings follow directly.
First, hierarchy is not a universal energy winner: if the application can tolerate learning from only the directly reachable subset, FedProx is the cheapest and often highly competitive option.
Second, once full-network participation matters, hierarchy becomes difficult to avoid under underwater reachability constraints.
This is the central systems insight of the paper.

\subsection{Energy Impact of Selective Cooperation and Compression}
\label{subsec:exp_cooperation}

The second question is whether fog-to-fog exchange should be activated continuously or only when the expected learning benefit justifies its cost.
The results show that always-on cooperation is rarely the most attractive option.
At $N{=}150$, HFL-Selective preserves the F1 score of HFL-Nearest while reducing energy by 33.3\%.
At $N{=}200$, the same saving is 31.4\%.
The remaining gap between HFL-Selective and HFL-NoCoop is modest relative to the cost of HFL-Nearest, which means selective cooperation behaves as intended: it only pays extra communication cost when cooperation may plausibly help smaller clusters.

A tier-wise breakdown clarifies where this gap comes from.
At $N{=}200$, HFL-NoCoop spends 31.3\,J on sensor-to-fog uploads and 102.7\,J on fog-to-gateway transmission.
HFL-Selective preserves these base terms and adds only 26.8\,J of fog-to-fog traffic, whereas HFL-Nearest adds 100.5\,J.
Thus, the penalty of always-on cooperation is driven almost entirely by unnecessary inter-fog exchange rather than by the underlying clustering path itself.

\begin{figure}[t]
\centering
\includegraphics[width=\linewidth]{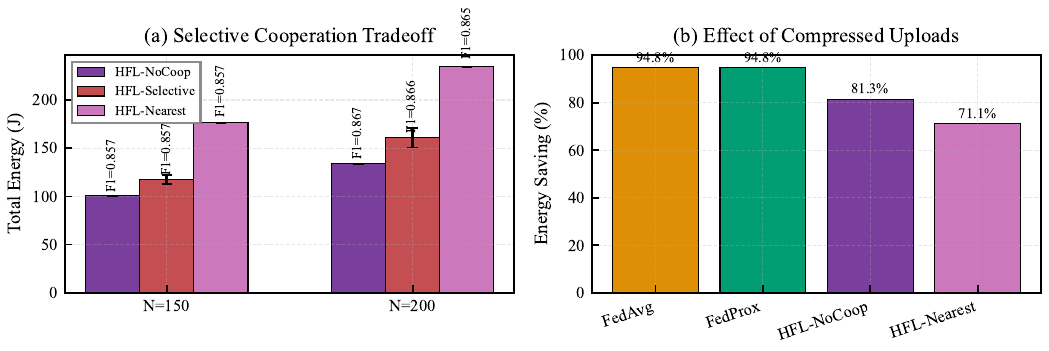}
\vspace{-5mm}
\caption{Engineering effects of selective cooperation and compressed uploads.
(a) Total energy of hierarchical methods at $N{=}150$ and $N{=}200$; the annotated F1-score values show that selective cooperation retains the accuracy of always-on cooperation while substantially reducing its cost.
(b) Compression savings in matched low-vs-full upload tests.
Error bars denote one standard deviation over three seeds where applicable.}
\label{fig:coop_compression}
\end{figure}

Compression is the dominant engineering lever across all evaluated methods.
In matched sensitivity tests with and without compressed uploads, moving from full-precision sensor uploads ($\rho{=}1.0$) to $\rho{=}0.05$ reduced total energy by 94.8\% for FedAvg and FedProx, 81.3\% for HFL-NoCoop, and 71.1\% for HFL-Nearest.
These gains are much larger than the difference between any two reasonable cooperation rules, so compressed uplinks are a prerequisite for practical underwater FL rather than a secondary optimisation.
The tier-level breakdown shows why compression and cooperation control are complementary: compression suppresses the sensor-upload term, while selective cooperation prevents the fog-to-fog term from dominating the hierarchical budget.

\subsection{Sensitivity to Data Heterogeneity}
\label{subsec:exp_noniid}

\begin{figure}[t]
\centering
\includegraphics[width=\linewidth]{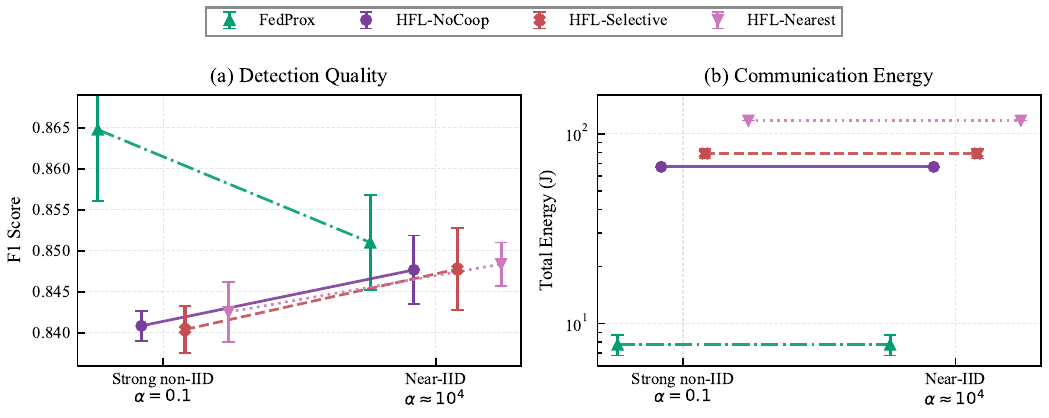}
\vspace{-5mm}
\caption{Non-IID sensitivity study at $N{=}100$, $M{=}10$, and $T{=}20$.
(a) F1 score under strongly non-IID and near-IID partitions.
(b) Total communication energy.
FedProx is strongest at $\alpha{=}0.1$ in this setting, but HFL-Selective remains close to the other hierarchical variants while using substantially less energy than HFL-Nearest.
Error bars denote one standard deviation over three seeds.}
\label{fig:noniid_sensitivity}
\end{figure}

To test whether the main conclusions remain consistent under heterogeneous client data, we ran an additional non-IID sensitivity study at $N{=}100$, $M{=}10$, and $T{=}20$ using Dirichlet partitions with $\alpha{=}0.1$ (strongly non-IID) and $\alpha{\approx}10^4$ (near-IID).
Figure~\ref{fig:noniid_sensitivity} compares FedProx, HFL-NoCoop, HFL-Selective, and HFL-Nearest under this setting.
First, hierarchy is not a universal remedy for data heterogeneity in this setting: at $\alpha{=}0.1$, FedProx achieves the highest F1 score (0.8648$\pm$0.0087), while the hierarchical methods cluster around 0.840--0.843.
Second, within the hierarchical family, HFL-Selective remains essentially tied with HFL-NoCoop and within one standard deviation of HFL-Nearest at both heterogeneity levels, while reducing the energy of always-on cooperation by about 33\%.
Thus, the non-IID result sharpens the main systems picture: selective cooperation is valuable because it preserves a favourable accuracy-energy balance within the hierarchical family, not because it guarantees the highest overall F1 in every regime.

\subsection{Real Benchmark Results}
\label{subsec:exp_real}

Finally, we evaluate the proposed framework on real anomaly-detection benchmarks to test whether the communication and learning trends observed in the synthetic setting also hold in practice.
We use three datasets: the Server Machine Dataset (SMD) and the Telemanom Soil Moisture Active Passive (SMAP) and Mars Science Laboratory (MSL) telemetry datasets.
Performance is reported with point-adjusted F1 (PA-F1), which is standard for segment-level anomaly evaluation but more generous than point-wise F1 because detecting any point within an anomalous segment credits the full segment.
SMD contains 10 machines with $D{=}38$ features, while SMAP contains 55 channels with $D{=}25$ features and MSL contains 27 channels with $D{=}55$.

\begin{figure}[t]
\centering
\includegraphics[width=\linewidth]{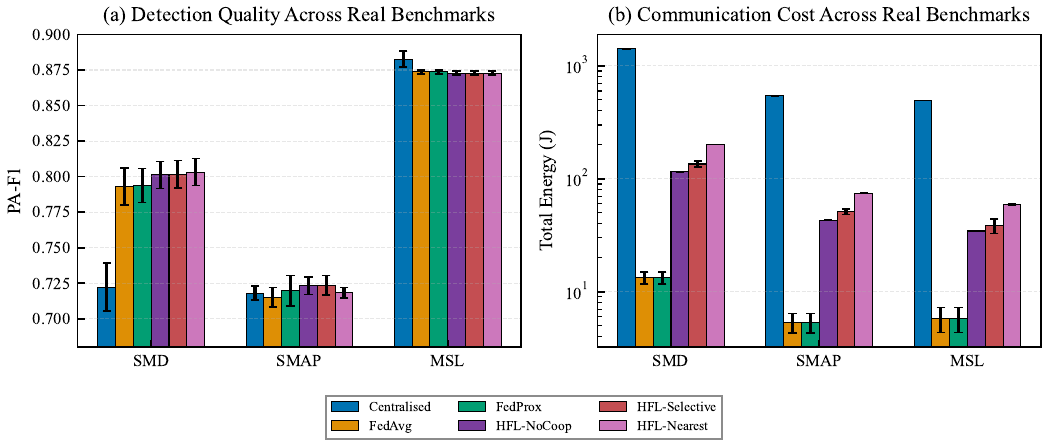}
\vspace{-5mm}
\caption{Grouped comparison across the SMD, SMAP, and MSL benchmarks.
(a) PA-F1 by method and dataset.
(b) Total communication energy on a logarithmic scale.
The figure makes the main real-data pattern visually explicit: flat FL offers the lowest-energy operating point, while low-overhead hierarchical methods remain competitive in detection quality and always-on cooperation is consistently the costliest hierarchical option.
Error bars denote one standard deviation over three seeds.}
\label{fig:real_tradeoff}
\end{figure}

Table~\ref{tab:telemanom_results} and Fig.~\ref{fig:real_tradeoff} show a clear and practically useful pattern.
On SMD, the hierarchical methods achieve the highest PA-F1, with HFL-NoCoop, HFL-Selective, and HFL-Nearest reaching 0.8013, 0.8015, and 0.8032, respectively, versus 0.7937 for FedProx and 0.7223 for the all-data centralised oracle.
On SMAP, FedProx, HFL-NoCoop, and HFL-Selective are tightly grouped around PA-F1 ${\approx}0.72$, with HFL-Selective highest at 0.7236 and the centralised oracle at 0.7181.
On MSL, the centralised oracle is strongest at 0.8827, while FedAvg/FedProx and the hierarchical methods remain close at 0.8727--0.8738.
Across all three real benchmarks, the same systems trend persists: flat FL methods define the lowest-energy operating point, low-overhead hierarchical methods remain highly competitive in detection quality, and always-on cooperation is the least attractive communication choice.
With only three seeds, the small PA-F1 gaps among FedProx, HFL-NoCoop, and HFL-Selective should be interpreted cautiously.

\begin{table*}[t]
\caption{Real-benchmark results over three seeds; performance is reported as PA-F1.}
\label{tab:telemanom_results}
\centering
\scriptsize
\begin{tabular}{@{}lcccccc@{}}
\toprule
\textbf{Method} & \textbf{SMD PA-F1} & \textbf{SMD E (J)} & \textbf{SMAP PA-F1} & \textbf{SMAP E (J)} & \textbf{MSL PA-F1} & \textbf{MSL E (J)} \\
\midrule
Centralised    & 0.7223$\pm$0.0168 & 1417.5$\pm$2.9 & 0.7181$\pm$0.0047 & 543.0$\pm$1.4 & 0.8827$\pm$0.0056 & 493.6$\pm$1.4 \\
FedAvg         & 0.7930$\pm$0.0130 &  13.3$\pm$1.6 & 0.7152$\pm$0.0068 &   5.4$\pm$1.1 & 0.8738$\pm$0.0015 &   5.8$\pm$1.5 \\
FedProx        & 0.7937$\pm$0.0119 &  13.3$\pm$1.6 & 0.7197$\pm$0.0106 &   5.4$\pm$1.1 & 0.8738$\pm$0.0015 &   5.8$\pm$1.5 \\
HFL-NoCoop     & 0.8013$\pm$0.0095 & 115.0$\pm$0.2 & 0.7233$\pm$0.0062 &  42.9$\pm$0.1 & 0.8727$\pm$0.0014 &  34.5$\pm$0.2 \\
HFL-Selective  & 0.8015$\pm$0.0097 & 135.2$\pm$8.1 & 0.7236$\pm$0.0067 &  51.3$\pm$3.0 & 0.8727$\pm$0.0014 &  38.5$\pm$5.6 \\
HFL-Nearest    & 0.8032$\pm$0.0095 & 201.7$\pm$0.2 & 0.7183$\pm$0.0037 &  74.6$\pm$0.2 & 0.8727$\pm$0.0014 &  59.2$\pm$0.9 \\
\bottomrule
\end{tabular}
\vspace{-2mm}
\end{table*}

\subsection{Discussion and Design Rules}
\label{subsec:discussion}

The paper's central contribution is to make acoustic feasibility explicit within a participation-aware evaluation framework, so that participation, energy, and learning quality can be assessed jointly under realistic underwater conditions.
The experiments yield four concrete IoUT design insights.

\emph{1) Participation must be reported alongside energy and accuracy.}
Under underwater reachability constraints, a flat method can be very efficient while excluding a large fraction of the network.
Accordingly, participation should be treated as a first-class evaluation metric rather than a supplementary diagnostic.
This also clarifies the energy trade-off: at $N{=}200$, FedProx uses about 0.16\,J per participating sensor, whereas HFL-NoCoop uses about 0.67\,J, so hierarchy is justified when coverage matters, not because it is universally more energy-efficient on a per-participant basis.

\emph{2) FedProx is the right flat baseline.}
Once feasible direct gateway links exist, FedProx is a very strong low-energy reference and cannot be dismissed by default.
Ignoring this baseline would overstate the value of hierarchy.

\emph{3) Always-on cooperation is wasteful.}
HFL-Nearest almost never buys enough extra accuracy to justify its fog-to-fog cost.
If hierarchy is required, HFL-NoCoop is the safest default and HFL-Selective is the principled way to add only limited cooperation.

\emph{4) Compression is indispensable.}
The largest reliable energy gains come from compressed sensor uploads, not from increasingly sophisticated coordination logic.
In practice, compression should be treated as mandatory infrastructure for IoUT FL.

% ------------------------------------------------------------------
\section{Conclusion}
\label{sec:conclusion}

This paper has presented an energy-efficient hierarchical federated learning framework for IoUT anomaly detection under realistic acoustic communication constraints.
The framework combines a feasibility-aware three-tier architecture, compressed sensor uploads, and selective cooperative aggregation among fog nodes, enabling detection quality, communication energy, and effective network participation to be evaluated within a single framework.

The experiments support three main conclusions.
First, flat FL and hierarchical FL answer different operational needs underwater: FedProx is the cheapest strong baseline, but at large scale it can train on only about 48\% of the network, whereas hierarchical methods preserve full participation through feasible fog paths.
Second, always-on inter-fog cooperation is unnecessary; HFL-Selective recovers the accuracy of HFL-Nearest while reducing its energy by 31--33\%, and HFL-NoCoop remains the strongest low-cost hierarchical default.
Third, compression is the dominant engineering gain, reducing total energy by 71--95\% in matched sensitivity tests across the evaluated methods.

Real-data experiments on SMD, SMAP, and MSL reinforce the same message.
Hierarchy is competitive in detection quality, but its value depends chiefly on whether the deployment prioritises full-network participation or minimum possible energy.
These results therefore provide concrete guidance for selecting between flat and hierarchical FL according to the operational priorities of a given IoUT deployment. 
Future work will extend the anomaly detector beyond point-wise autoencoders, study asynchronous/mobile fog settings, and investigate adaptive decision rules, including learning-based orchestrators, that can generalise across different sensor/fog counts without retraining.
Code, scripts, and generated result files will be made available to reviewers on request and released publicly upon acceptance.

\section*{Acknowledgment}
The authors used OpenAI GPT-based tools for language refinement and limited prose revision in the Abstract, Introduction, Related Work, Evaluation, Discussion, and Conclusion; all scientific content, code, results, and final wording were verified by the authors.

\bibliographystyle{IEEEtran}
\bibliography{References_iot}

@article{qiu2019underwater,
  title={Internet of underwater things: Architecture, protocols, and applications},
  author={Qiu, T. and Zhao, Z. and Zhang, T. and Chen, C. and Zhou, P.},
  journal={IEEE Internet of Things Journal},
  volume={7},
  number={8},
  pages={6671--6689},
  year={2019},
  publisher={IEEE}
}

@article{stojanovic2009underwater,
  title={Underwater acoustic communication channels: Propagation models and statistical characterization},
  author={Stojanovic, M. and Preisig, J.},
  journal={IEEE Communications Magazine},
  volume={47},
  number={1},
  pages={84--89},
  year={2009},
  publisher={IEEE}
}

@article{jouhari2019underwater,
  title={Underwater wireless sensor networks: A survey on enabling technologies, localization protocols, and internet of underwater things},
  author={Jouhari, M. and Ibrahimi, K. and Tembine, H. and Ben-Othman, J.},
  journal={IEEE Access},
  volume={7},
  pages={96879--96899},
  year={2019},
  publisher={IEEE}
}

@article{akyildiz2005underwater,
  title={Underwater acoustic sensor networks: research challenges},
  author={Akyildiz, I. F. and Pompili, D. and Melodia, T.},
  journal={Ad hoc networks},
  volume={3},
  number={3},
  pages={257--279},
  year={2005},
  publisher={Elsevier}
}

@inproceedings{mcmahan2017communication,
  title={Communication-efficient learning of deep networks from decentralized data},
  author={McMahan, B. and Moore, E. and Ramage, D. and Hampson, S. and Arcas, B. A.},
  booktitle={Artificial intelligence and statistics},
  pages={1273--1282},
  year={2017},
  organization={PMLR}
}

@inproceedings{liu2020client,
  title={Client-Edge-Cloud Hierarchical Federated Learning},
  author={Liu, Lumin and Zhang, Jun and Song, Shenghui and Letaief, Khaled B.},
  booktitle={2020 IEEE International Conference on Communications (ICC)},
  pages={1--6},
  year={2020},
  organization={IEEE},
  doi={10.1109/ICC40277.2020.9148862}
}

@article{wang2019adaptive,
  title={Adaptive federated learning in resource constrained edge computing systems},
  author={Wang, S. and Tuor, T. and Salonidis, T. and Leung, K. K. and Makaya, C.},
  journal={IEEE Journal on Selected Areas in Communications},
  volume={37},
  number={6},
  pages={1205--1221},
  year={2019},
  publisher={IEEE}
}

@article{yang2020energy,
  title={Energy-efficient federated learning over wireless communication networks},
  author={Yang, Z. and Chen, M. and Saad, W. and Hong, C. S. and Shikh-Bahaei, M.},
  journal={IEEE Transactions on Wireless Communications},
  volume={20},
  number={3},
  pages={1935--1949},
  year={2020},
  publisher={IEEE}
}

@inproceedings{abad2019hflcellular,
  title     = {Hierarchical Federated Learning Across Heterogeneous Cellular Networks},
  author    = {Abad, Mehdi Salehi Heydar and Ozfatura, Emre and G{\"u}nd{\"u}z, Deniz and Er{\c c}etin, {\"O}zg{\"u}r},
  booktitle = {2020 IEEE International Conference on Acoustics, Speech and Signal Processing (ICASSP)},
  pages     = {8866--8870},
  year      = {2020},
  organization = {IEEE},
  doi       = {10.1109/ICASSP40776.2020.9054634}
}

@article{ghosh2020ifca,
  title   = {An Efficient Framework for Clustered Federated Learning},
  author  = {Ghosh, Avishek and Chung, Jichan and Yin, Dong and Ramchandran, Kannan},
  journal = {IEEE Transactions on Information Theory},
  volume  = {68},
  number  = {12},
  pages   = {8076--8091},
  year    = {2022},
  doi     = {10.1109/TIT.2022.3192506}
}

@inproceedings{gong2019memae,
  title     = {Memorizing Normality to Detect Anomaly: Memory-Augmented Deep Autoencoder for Unsupervised Anomaly Detection},
  author    = {Gong, Dong and Liu, Lingqiao and Le, Vuong and Saha, Budhaditya and Mansour, Moussa Reda and Venkatesh, Svetha and van den Hengel, Anton},
  booktitle = {Proceedings of the IEEE/CVF International Conference on Computer Vision (ICCV)},
  year      = {2019}
}

@inproceedings{karimireddy2020scaffold,
  title     = {{SCAFFOLD}: Stochastic Controlled Averaging for Federated Learning},
  author    = {Karimireddy, Sai Praneeth and Kale, Satyen and Mohri, Mehryar and Reddi, Sashank J. and Stich, Sebastian U. and Suresh, Ananda Theertha},
  booktitle = {Proceedings of the 37th International Conference on Machine Learning},
  series    = {Proceedings of Machine Learning Research},
  volume    = {119},
  pages     = {5132--5143},
  year      = {2020},
  publisher = {PMLR}
}

@inproceedings{li2020fedprox,
  title     = {Federated Optimization in Heterogeneous Networks},
  author    = {Li, Tian and Sahu, Anit Kumar and Zaheer, Manzil and Sanjabi, Maziar and Talwalkar, Ameet and Smith, Virginia},
  booktitle = {Proceedings of Machine Learning and Systems},
  volume    = {2},
  pages     = {429--450},
  year      = {2020},
  publisher = {MLSys}
}

@inproceedings{lin2018dgc,
  title     = {Deep Gradient Compression: Reducing the Communication Bandwidth for Distributed Training},
  author    = {Lin, Yujun and Han, Song and Mao, Huizi and Wang, Yu and Dally, William J.},
  booktitle = {International Conference on Learning Representations (ICLR)},
  year      = {2018},
  eprint    = {1712.01887},
  archivePrefix = {arXiv},
  primaryClass  = {cs.LG}
}

@inproceedings{malhotra2016lstmencdec,
  title     = {Long Short Term Memory Networks for Anomaly Detection in Time Series},
  author    = {Malhotra, Pankaj and Vig, Lovekesh and Shroff, Gautam and Agarwal, Puneet},
  booktitle = {23rd European Symposium on Artificial Neural Networks, Computational Intelligence and Machine Learning (ESANN)},
  year      = {2015},
  url       = {https://www.esann.org/sites/default/files/proceedings/legacy/es2015-56.pdf}
}

@article{nguyen2021fliotsurvey,
  title   = {Federated Learning for Internet of Things: A Comprehensive Survey},
  author  = {Nguyen, Dinh C. and Ding, Ming and Pathirana, Pubudu N. and Seneviratne, Aruna and Li, Jun and Poor, H. Vincent},
  journal = {IEEE Communications Surveys \& Tutorials},
  volume  = {23},
  number  = {3},
  pages   = {1622--1658},
  year    = {2021},
  doi     = {10.1109/COMST.2021.3075439}
}

@inproceedings{ruff2018deepsvdd,
  title     = {Deep One-Class Classification},
  author    = {Ruff, Lukas and Vandermeulen, Robert A. and G{\"o}rnitz, Nico and Deecke, Lucas and Siddiqui, Shoaib A. and Binder, Alexander and M{\"u}ller, Emmanuel and Kloft, Marius},
  booktitle = {Proceedings of the 35th International Conference on Machine Learning},
  series    = {Proceedings of Machine Learning Research},
  volume    = {80},
  pages     = {4393--4402},
  year      = {2018},
  publisher = {PMLR}
}

@book{urick1983principles,
  title        = {Principles of Underwater Sound},
  author       = {Urick, Robert J.},
  edition      = {3},
  year         = {1983},
  publisher    = {McGraw-Hill},
  address      = {New York, NY, USA}
}

@article{thorp1967analytical,
  title   = {Analytic Description of the Low-Frequency Attenuation Coefficient},
  author  = {Thorp, W. H.},
  journal = {The Journal of the Acoustical Society of America},
  volume  = {42},
  number  = {1},
  pages   = {270--270},
  year    = {1967}
}

@article{wenz1962acoustic,
  title   = {Acoustic Ambient Noise in the Ocean: Spectra and Sources},
  author  = {Wenz, G. M.},
  journal = {The Journal of the Acoustical Society of America},
  volume  = {34},
  number  = {12},
  pages   = {1936--1956},
  year    = {1962}
}

@inproceedings{stojanovic2007underwater,
  title     = {Underwater Acoustic Communications: Design Considerations on the Physical Layer},
  author    = {Stojanovic, Milica},
  booktitle = {Proceedings of the 5th Annual Conference on Wireless On-demand Network Systems and Services (WONS)},
  pages     = {1--10},
  year      = {2007},
  organization = {IEEE}
}

@article{chitre2008underwater,
  title   = {Underwater Acoustic Communications and Networking: Recent Advances and Future Challenges},
  author  = {Chitre, Mandar and Shahabudeen, Shiraz and Stojanovic, Milica},
  journal = {Marine Technology Society Journal},
  volume  = {42},
  number  = {1},
  pages   = {103--116},
  year    = {2008}
}

@article{adam2024iout_security,
  title   = {State-of-the-Art Security Schemes for the {Internet of Underwater Things}: A Holistic Survey},
  author  = {Adam, Noha and Ali, Muhammad and Naeem, Faisal and Ghazy, Ahmed S. and Kaddoum, Georges},
  journal = {IEEE Open Journal of the Communications Society},
  volume  = {5},
  pages   = {6561--6590},
  year    = {2024},
  doi     = {10.1109/OJCOMS.2024.3478035}
}

@article{stewart2025iout_oil_gas,
  title   = {Towards the Underwater {Internet of Things} for Subsea Oil and Gas Monitoring},
  author  = {Stewart, Craig and Ayaz, Basit and Fough, Neda and Prabhu, Radhakrishna},
  journal = {Internet of Things},
  volume  = {30},
  pages   = {101510},
  year    = {2025},
  doi     = {10.1016/j.iot.2025.101510}
}

@article{zeng2020survey_uwan,
  title   = {A Survey of Underwater Acoustic Networks for the {Internet of Underwater Things}},
  author  = {Zeng, Zhiqiang and Fu, Shu and Zhang, Huahui and Dong, Yabo},
  journal = {IEEE Communications Surveys \& Tutorials},
  volume  = {22},
  number  = {2},
  pages   = {908--944},
  year    = {2020},
  doi     = {10.1109/COMST.2020.2965924}
}

@article{pei2023fediout,
  title   = {{Fed-IoUT}: Opportunities and Challenges of Federated Learning in the {Internet of Underwater Things}},
  author  = {Pei, Jiacheng and Liu, Wei and Wang, Lei and Liu, Chen and Bashir, Ali Kashif and Wang, Yinan},
  journal = {IEEE Internet of Things Magazine},
  volume  = {6},
  number  = {1},
  pages   = {108--112},
  year    = {2023},
  doi     = {10.1109/IOTM.001.2200188}
}

@article{popli2025fl_underwater_drones,
  title   = {A Federated Learning Framework for Enhanced Data Security and Cyber Intrusion Detection in Distributed Network of Underwater Drones},
  author  = {Popli, Manpreet Singh and Singh, Rajesh Pal and Popli, Navjot Kaur and Mamun, Md. Al},
  journal = {IEEE Access},
  volume  = {13},
  pages   = {12634--12646},
  year    = {2025},
  doi     = {10.1109/ACCESS.2025.3530499}
}

@incollection{shaheen2024fl_iout,
  title     = {Revolutionizing {Internet of Underwater Things} with Federated Learning},
  author    = {Shaheen, Momina and Farooq, Muhammad Shoaib and Umer, Tariq and Tran, Thanh Anh},
  booktitle = {Artificial Intelligence and Edge Computing for Sustainable Ocean Health},
  series    = {Springer Series in Applied Machine Learning},
  publisher = {Springer},
  year      = {2024},
  doi       = {10.1007/978-3-031-64642-3_12}
}

@article{luo2020hfel_twc,
  title   = {{HFEL}: Joint Edge Association and Resource Allocation for Cost-Efficient Hierarchical Federated Edge Learning},
  author  = {Luo, Siqi and Chen, Xu and Wu, Qiong and Zhou, Zhi and Yu, Shuai},
  journal = {IEEE Transactions on Wireless Communications},
  volume  = {19},
  number  = {10},
  pages   = {6535--6548},
  year    = {2020},
  doi     = {10.1109/TWC.2020.3003744}
}

@article{wang2022feduc,
  title   = {{FedUC}: A Unified Clustering Approach for Hierarchical Federated Learning},
  author  = {Wang, Zhiqiang and others},
  journal = {IEEE Transactions on Mobile Computing},
  year    = {2024},
  volume  = {23},
  number  = {10},
  pages   = {9614--9630},
  doi     = {10.1109/TMC.2024.3366947}
}

@article{mhaisen2022hfl_topology,
  title   = {Optimal User-Edge Assignment in Hierarchical Federated Learning Based on Statistical Properties and Network Topology Constraints},
  author  = {Mhaisen, Naram and Awad, Ali and Mohamed, Amr and Erbad, Aiman and Guizani, Mohsen},
  journal = {IEEE Transactions on Network Science and Engineering},
  volume  = {9},
  number  = {1},
  pages   = {55--66},
  year    = {2022},
  doi     = {10.1109/TNSE.2021.3053588}
}

@article{liu2023async_fl_mobile,
  title   = {Adaptive Asynchronous Federated Learning in Resource-Constrained Edge Computing},
  author  = {Liu, Jianchun and Xu, Hongli and Wang, Lun and Xu, Yang and Qian, Chen and Huang, Jingyang and Huang, He},
  journal = {IEEE Transactions on Mobile Computing},
  volume  = {22},
  number  = {2},
  pages   = {674--690},
  year    = {2023},
  doi     = {10.1109/TMC.2021.3098182}
}

@article{li2020fl_challenges_survey,
  title   = {Federated Learning: Challenges, Methods, and Future Directions},
  author  = {Li, Tian and Sahu, Anit Kumar and Talwalkar, Ameet and Smith, Virginia},
  journal = {IEEE Signal Processing Magazine},
  volume  = {37},
  number  = {3},
  pages   = {50--60},
  year    = {2020},
  doi     = {10.1109/MSP.2020.2975749}
}

@article{khan2021fl_iot_survey,
  title   = {Federated Learning for {Internet of Things}: Recent Advances, Taxonomy, and Open Challenges},
  author  = {Khan, Latif U. and Saad, Walid and Han, Zhu and Hossain, Ekram and Hong, Choong Seon},
  journal = {IEEE Communications Surveys \& Tutorials},
  volume  = {23},
  number  = {3},
  pages   = {1759--1799},
  year    = {2021},
  doi     = {10.1109/COMST.2021.3090430}
}

@inproceedings{reddi2021adaptive_fl,
  title   = {Adaptive Federated Optimization},
  author  = {Reddi, Sashank and Charles, Zachary and Zaheer, Manzil and Garrett, Zachary and Rush, Keith and Kone\v{c}n\'{y}, Jakub and Kumar, Sanjiv and McMahan, H. Brendan},
  booktitle = {International Conference on Learning Representations (ICLR)},
  year    = {2021},
  url     = {https://openreview.net/forum?id=LkFG3lB13U5}
}

@inproceedings{lian2017decentralized_sgd,
  title   = {Can Decentralized Algorithms Outperform Centralized Algorithms? {A} Case Study for Decentralized Parallel Stochastic Gradient Descent},
  author  = {Lian, Xiangru and Zhang, Ce and Zhang, Huan and Hsieh, Cho-Jui and Zhang, Wei and Liu, Ji},
  booktitle = {Advances in Neural Information Processing Systems (NeurIPS)},
  volume  = {30},
  year    = {2017}
}

@article{chen2021energy_fl_convergence,
  title   = {A Joint Learning and Communications Framework for Federated Learning Over Wireless Networks},
  author  = {Chen, Mingzhe and Yang, Zhaohui and Saad, Walid and Yin, Changchuan and Poor, H. Vincent and Cui, Shuguang},
  journal = {IEEE Transactions on Wireless Communications},
  volume  = {20},
  number  = {1},
  pages   = {269--283},
  year    = {2021},
  doi     = {10.1109/TWC.2020.3024629}
}

@inproceedings{wang2020fedrl_noniid,
  title     = {Optimizing Federated Learning on Non-{IID} Data with Reinforcement Learning},
  author    = {Wang, Hao and Kaplan, Zakhary and Niu, Di and Li, Baochun},
  booktitle = {Proceedings of IEEE INFOCOM},
  pages     = {1698--1707},
  year      = {2020},
  doi       = {10.1109/INFOCOM41043.2020.9155494}
}

@inproceedings{he2024fedaa_rl_agg,
  title     = {{FedAA}: A Reinforcement Learning Perspective on Adaptive Aggregation for Fair and Robust Federated Learning},
  author    = {He, Jialuo and Chen, Wei and Zhang, Xiaojin},
  booktitle = {Proceedings of the AAAI Conference on Artificial Intelligence},
  volume    = {39},
  number    = {16},
  pages     = {17085--17093},
  year      = {2025},
  doi       = {10.1609/AAAI.V39I16.33878}
}

@article{hu2020rl_uwsn_routing,
  title   = {Reinforcement Learning Based Routing Protocol for Underwater Acoustic Sensor Networks},
  author  = {Hu, Tangxin and Fei, Yuntao},
  journal = {IEEE Internet of Things Journal},
  volume  = {7},
  number  = {11},
  pages   = {11215--11226},
  year    = {2020},
  doi     = {10.1109/JIOT.2020.2990975}
}

@article{jiang2023drl_uwsn_routing,
  title   = {Deep Reinforcement Learning-Based Multi-Hop Routing for Underwater Wireless Sensor Networks},
  author  = {Jiang, Peng and others},
  journal = {IEEE Internet of Things Journal},
  volume  = {10},
  number  = {22},
  pages   = {19731--19744},
  year    = {2023},
  doi     = {10.1109/JIOT.2023.3283018}
}

@article{omeke2021dekcs,
  title   = {{DEKCS}: A Dynamic Clustering Protocol to Prolong Underwater Sensor Networks},
  author  = {Omeke, Kenechi G. and Mollel, Michael S. and Ozturk, Metin and Ansari, Shuja and Zhang, Lei and Abbasi, Qammer H. and Imran, Muhammad Ali},
  journal = {IEEE Sensors Journal},
  volume  = {21},
  number  = {7},
  pages   = {9457--9464},
  year    = {2021},
  doi     = {10.1109/JSEN.2021.3054943}
}

@article{pang2021deep_ad_survey,
  title   = {Deep Learning for Anomaly Detection: A Review},
  author  = {Pang, Guansong and Shen, Chunhua and Cao, Longbing and van den Hengel, Anton},
  journal = {ACM Computing Surveys},
  volume  = {54},
  number  = {2},
  pages   = {1--38},
  year    = {2021},
  doi     = {10.1145/3439950}
}

@article{ruff2021unifying_ad,
  title   = {A Unifying Review of Deep and Shallow Anomaly Detection},
  author  = {Ruff, Lukas and Kauffmann, Jacob R. and Vandermeulen, Robert A. and Montavon, Gr\'{e}goire and Samek, Wojciech and Kloft, Marius and Dietterich, Thomas G. and M\"{u}ller, Klaus-Robert},
  journal = {Proceedings of the IEEE},
  volume  = {109},
  number  = {5},
  pages   = {756--795},
  year    = {2021},
  doi     = {10.1109/JPROC.2021.3052449}
}

@article{liu2022intrusion_maritime,
  title   = {Intrusion Detection for Maritime Transportation Systems with Batch Federated Aggregation},
  author  = {Liu, Weiyang and Xu, Xiaolong and Wu, Li and Qi, Lianyong and Jolfaei, Alireza and Ding, Weiping and Khosravi, Mohammad R.},
  journal = {IEEE Transactions on Intelligent Transportation Systems},
  volume  = {24},
  number  = {2},
  pages   = {2503--2514},
  year    = {2022},
  doi     = {10.1109/TITS.2022.3223145}
}

@article{aouedi2022fed_semisup_iiot,
  title   = {Federated Semisupervised Learning for Attack Detection in Industrial {Internet of Things}},
  author  = {Aouedi, Ons and Piamrat, Kandaraj and Muller, Guillaume and Singh, Kamal},
  journal = {IEEE Transactions on Industrial Informatics},
  volume  = {19},
  number  = {1},
  pages   = {286--295},
  year    = {2022},
  doi     = {10.1109/TII.2022.3156642}
}

@article{xiao2020edge_iout,
  title   = {Edge Computing for the {Internet of Underwater Things}: A Survey},
  author  = {Xiao, Lianfang and Wan, Xiaoran and Lu, Xiaojun and Zhang, Yanan and Wu, Di},
  journal = {IEEE Communications Surveys \& Tutorials},
  volume  = {22},
  number  = {2},
  pages   = {1087--1112},
  year    = {2020},
  doi     = {10.1109/COMST.2020.2970701}
}

@inproceedings{stich2018sparsified,
  title     = {Sparsified {SGD} with Memory},
  author    = {Stich, Sebastian U. and Cordonnier, Jean-Baptiste and Jaggi, Martin},
  booktitle = {Advances in Neural Information Processing Systems (NeurIPS)},
  volume    = {31},
  year      = {2018}
}

@inproceedings{alistarh2018sparsified,
  title     = {The Convergence of Sparsified Gradient Methods},
  author    = {Alistarh, Dan and Hoefler, Torsten and Johansson, Mikael and Konstantinov, Nikola and Khirirat, Sarit and Renggli, C{\'e}dric},
  booktitle = {Advances in Neural Information Processing Systems (NeurIPS)},
  volume    = {31},
  year      = {2018}
}

@inproceedings{dettmers2016bit,
  title     = {8-Bit Approximations for Parallelism in Deep Learning},
  author    = {Dettmers, Tim},
  booktitle = {International Conference on Learning Representations (ICLR), Workshop Track},
  year      = {2016}
}

\end{document}